\documentclass[a4paper, 10pt, conference]{ieeeconf}      

\usepackage[dvipsnames]{xcolor}
\usepackage{colortbl}
\let\labelindent\relax
\usepackage{enumitem}
\usepackage{subfigure}
\usepackage{FG2026}

\definecolor{cvprblue}{rgb}{0.21,0.49,0.74}
\definecolor{colorserena}{rgb}{0.99,0.1,0.5}
\definecolor{colorbertrand}{rgb}{0.0,0.99,0.0}
\usepackage[pagebackref,breaklinks,colorlinks,allcolors=cvprblue]{hyperref}
\usepackage{amssymb}
\usepackage{pifont}
\usepackage{booktabs}
\usepackage{threeparttable}
\usepackage{makecell}
\usepackage{multicol}
\usepackage{multirow}
\usepackage{graphicx}
\usepackage{rotating}
\usepackage{mathtools}
\usepackage[ruled,vlined]{algorithm2e}
\usepackage{amsmath,amssymb}
\usepackage{float}
\newcommand{\cross}[1][1pt]{\ooalign{%
  \rule[1ex]{1ex}{#1}\cr
  \hss\rule{#1}{.7em}\hss\cr}}
\DeclarePairedDelimiter\ceil{\lceil}{\rceil}

\usepackage{tablefootnote}
\newcommand{\shorteq}{\mathrel{\mkern0.2mu\mathpalette\shorteq@\relax\mkern0.2mu}}
\newcommand{\cmark}{\textcolor{LimeGreen}{\ding{51}}}%
\newcommand{\xmark}{\textcolor{BrickRed}{\ding{55}}}%
\newcommand{\HIDEINSTRUCTIONS}[1]{}
\let\titleold\title
\renewcommand{\title}[1]{\titleold{#1}\newcommand{\thetitle}{#1}}
\def\maketitlesupplementary
   {
   \newpage
       \twocolumn[
        \centering
        \Large
        \textbf{\thetitle}\\
        \vspace{0.5em}Supplementary Material \\
        \vspace{1.0em}
       ] 
   }

\FGfinalcopy 

\IEEEoverridecommandlockouts                              
\def\FGPaperID{374} 

\title{\LARGE \bf
 HUI360 : A 360° Egocentric Dataset and Baselines for Human-Robot Interaction Anticipation 
}

\author{\parbox{16cm}{\centering
    {\large Raphael Lorenzo-Louis$^{1,2}$, Fabio Amadio$^1$, Bertrand Luvison$^2$, and Serena Ivaldi$^1$}\\
    {\normalsize
    $^1$ Inria, CNRS, UL, Loria, HUCEBOT, F-54600 Villers-les-Nancy, France\\
    $^2$ Universit\'e Paris-Saclay, CEA, List, F-91120 Palaiseau, France}}
}

\usepackage{fancyhdr}

\begin{document}

\ifFGfinal
\thispagestyle{empty}
\pagestyle{empty}
\else
\author{Anonymous FG2026 submission\\ Paper ID \FGPaperID \\}
\pagestyle{plain}
\fi

\maketitle

\thispagestyle{fancy}
\renewcommand{\headrulewidth}{0pt}
\fancyhf{}
\fancyhead[C]{2026 International Conference on Automatic Face and Gesture Recognition (FG)}
\fancyfoot[L]{979-8-3315-7231-0/26/\$31.00\ \copyright\ 2026\ European\ Union}

\begin{figure*}[htbp]
  \centering
  \includegraphics[width=\linewidth]{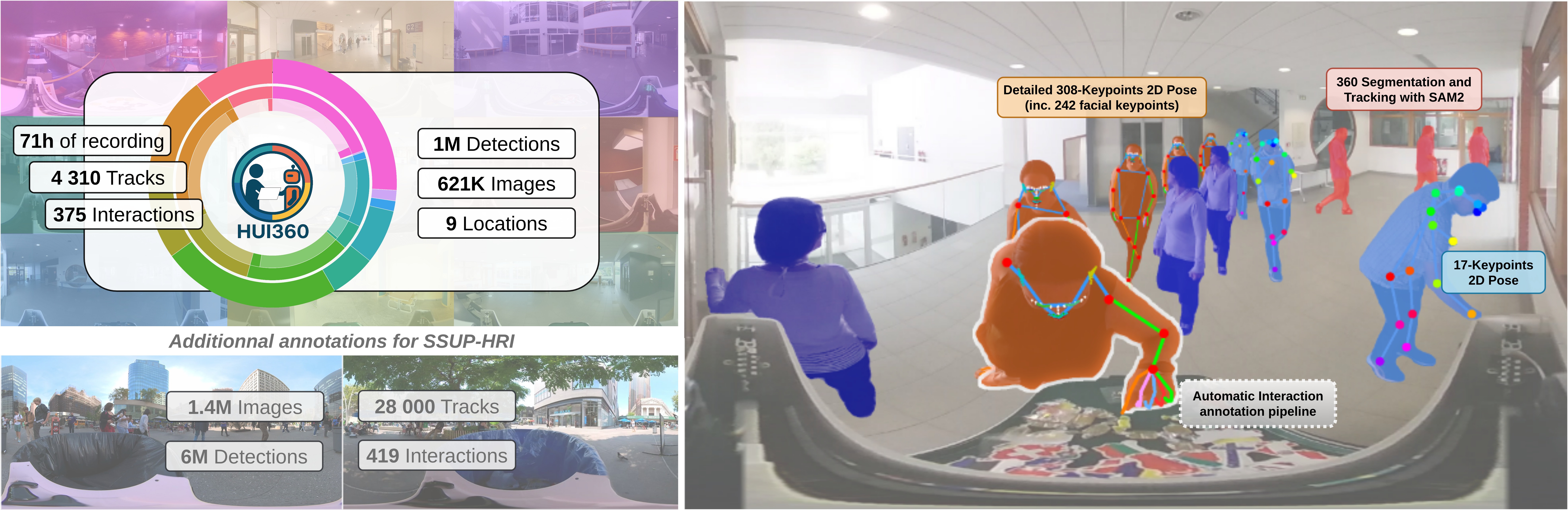}
  \caption{%
      Overview of the content of HUI360. The dataset comprises 4 310 tracks forming more than 1M detections over time, for each of them we provide a segmentation mask as well as detailed 2D Pose Keypoints, images are available on request for research purposes. The dataset is diverse (9 environments) and interaction are automatically annotated. Automatic tracking and interaction annotation is manually reviewed and corrected. Additional annotations enabling cross-dataset studies have been generated from the data of SSUP-HRI \cite{bu2024ssuphrisocialsignalingurban} using our automated pipeline.%
  }
  \label{fig:concept}
\end{figure*}

\begin{abstract}

As robots increasingly operate in human-populated environments, anticipating human intentions is essential for enabling proactive and socially aware behavior. Automatic anticipation of human–robot interactions is thus emerging as a crucial perception challenge for embodied agents. 

To this end, we introduce HUI360, the largest dataset for human-robot interaction anticipation in the wild and its set of baselines. 
The dataset was collected from a mobile robot, in the wild, over multiple days within a 3-month period, and in several environments, capturing natural, spontaneous behaviors from both passersby and users, and encompassing a diverse range of individuals.
This variety enables evaluating and improving the generalization capabilities of interaction anticipation models.

We designed a pipeline and share code for automatic interaction annotation in arbitrary 360° equirectangular videos, along with interfaces for manual refinement.
Using this pipeline, we release the HUI360 open set of 1M pre-processed annotations, including detailed 2D poses, facial keypoints, and segmentation masks, obtained using state-of-the-art computer vision methods and manually curated to ensure high-quality tracking and interaction annotation. 
Additionally, we release the raw panoptic 360° images captured from the robot’s egocentric viewpoint (on demand, for research purpose only in compliance with GDPR). 

Finally, we establish benchmark baselines for interaction anticipation, including the first cross-dataset evaluations for this task: to this end, we also release 6M annotations for another existing in-the-wild outdoor dataset collected from a mobile robot (SSUP-HRI).

Dataset and code can be found at \url{https://hucebot.github.io/hui360}.

\end{abstract}

\section{INTRODUCTION}


To better assist humans in public and private places, robots need the ability to predict if a passerby is interested in the robot and is going to interact with it. This ability allow personal and service robots to proactively engage the interaction with the human, yielding a better user experience and a more efficient execution of their mission. 

In the Human-Robot Interaction (HRI) literature, intent prediction definitions are often fuzzy and depend on the context of the interaction~\cite{pard_thompson,Abbate_2024,bian2025interactmejointegocentric,mohsen2025mintrvaemulticuesintentionprediction}.
It is also difficult to find a unique definition of ``what is an interaction'' and ``what is intent'': is it established at the mutual gaze phase, when humans have approached the robot, or when humans have touched or physically interacted with the robot?

In this paper, we alleviate this unclarity and formulate the problem as \textit{interaction anticipation}, leveraging non-ambiguous physical interactions, like picking and placing.
We do not consider the many cases where people stare at the robot, with curiosity or amusement. In fact, it is difficult to determine objectively the onset of these interactions, making labeling very fuzzy and subject to interpretation.

Intention and engagement in interaction are linked to the dynamics of verbal and nonverbal cues, such as utterances, gaze and posture \cite{Anzalone2015,ivaldi2015engagementmodelsconsiderindividual}; but in social interaction scenarios in the wild, where many possible human passerby can be potentially interact with the robot, there are other dynamics and factors at play, and these cues are not only difficult to extract from videos, but may be insufficient to provide a robust prediction.

In this case, to provide reliable predictions, data-driven approaches are more convenient, but they rely on annotated datasets that are lacking in the literature. Some exist, such as SSUP-HRI \cite{bu2024ssuphrisocialsignalingurban} or Shutter \cite{pard_thompson} (see Table~\ref{tab:comparison} for an in-depth analysis), but they are often too small for training \cite{mohsen2025mintrvaemulticuesintentionprediction}, lack diversity in context \cite{benyoussef:hal-02943475}, use specific hardware and software that limits extendability \cite{pard_thompson}, or do not capture humans behaving spontaneously and naturally in the wild \cite{Vaufreydaz_2016}.
Furthermore, another fundamental problem in the literature is the lack of reproducible and comparable references, as well as domain transfer evaluations.

To address these issues, our contributions are: \textbf{(1)} the largest dataset for human-robot interaction anticipation in the wild -- HUI360; \textbf{(2)} an open-source method to detect, track and automatically annotate human-robot interactions in any 360 video stream to encourage dataset extendability and reproducibility; 
\textbf{(3)}, a formalized evaluation protocol for the anticipation measure and associated baselines, including cross-location and cross-dataset evaluations that demonstrates the importance of a large and diverse dataset for this task.

\label{sec:intro}

\section{RELATED WORKS}
\label{sec:related}
\begin{table*}[ht]
\centering
\caption{Comparison of Human-Robot Interaction Datasets. \textbf{\#Frames} and \textbf{FOV} are either: reported from the original publications, reported from the downloaded dataset when available or estimated using reported durations and framerate. We added the unlabeled dataset of \cite{bu2024ssuphrisocialsignalingurban} with numbers reported from the original publication about the first of their two field session. We denote \textbf{SSUP-A} the SSUP-HRI data from both field sessions augmented with our annotations. \textbf{Available} datasets are either public \cite{pard_thompson, BrscicKanda2013,mohsen2025mintrvaemulticuesintentionprediction,ryoo2015prediction} or can be obtained via a defined procedure \cite{benyoussef:hal-02943475,bu2024ssuphrisocialsignalingurban}.}
\fontsize{8}{10}\selectfont
\begin{tabular}{lcccccccccc}
\toprule
 & \textbf{\#Frames} & \textbf{\#Tracks} & \textbf{\#Interaction} & \textbf{Available} & \textbf{Curated} & \textbf{FOV} & \textbf{Poses} & \textbf{Video} & \textbf{\#Scenes} & \textbf{ITW} \\
\midrule
\textbf{SSUP-HRI \cite{bu2024ssuphrisocialsignalingurban}} & 1.4M & 0 & 118 & \cmark & \xmark & 360 & \xmark & \cmark & Mobile & \cmark \\
\textbf{JPL-Int.-Ext. \cite{ryoo2015prediction}} & 21K & 180 & 180 & \cmark & \cmark & 90 & Kinect & \cmark & Mobile & \xmark \\
\textbf{ATC-Aug. \cite{pard_thompson,BrscicKanda2013}} & 33K & 4 245 & 112 & \cmark & \cmark & 360 & \xmark & \xmark & Mobile & \cmark \\
\textbf{Abbate et al. \cite{Abbate_2024}} & Unk. & 3 422 & Unk. & \xmark & \xmark & 90 & Kinect & \xmark & 3 & \cmark \\
\textbf{Shutter \cite{pard_thompson}} & 58K & 1 057 & 314 & \cmark & \xmark & 100 & Kinect & \xmark & 2 & \cmark \\
\textbf{Vaufreydaz et al. \cite{Vaufreydaz_2016}} & 180K & 29 & 29 & \xmark & \xmark & 90 & Kinect & \xmark & 2 & \xmark \\
\textbf{UE-HRI \cite{benyoussef:hal-02943475}} & 207K & 195\tablefootnote{54 out of the 195 are shared.} & 195 & \cmark & \cmark & 56 & \xmark & \cmark & 1 & \cmark \\
\textbf{MINT-RVAE \cite{mohsen2025mintrvaemulticuesintentionprediction}} & 12K & 88 & 88 & \cmark & \cmark & Unk. & 2D & \xmark & 3 & \xmark \\
\midrule
\textbf{Ours} & 612K & 4310 & 375 & \cmark & \cmark & 360 & 2D & \cmark & 9 & \cmark \\
\midrule
\textbf{Ours+SSUP-A} & 1.8M\tablefootnote{We processed the data of SSUP-HRI at 15fps instead of the original 30fps and manually filtered out some invalid parts,  resulting in 1.15M frames in \textbf{SSUP-A} out of the 9.8h and 12.1h of recordings of the first and second field session of \cite{bu2024ssuphrisocialsignalingurban}.} & 32 008 & 794 & \cmark & \cmark & 360 & 2D & \cmark & 9 + Mobile & \cmark \\
\bottomrule
\end{tabular}

\label{tab:comparison}
\end{table*}

\subsection{Engagement and disengagement assessment}

The initiation of an interaction is closely related to user engagement, a notoriously vague \cite{Sorrentino24} yet widely used notion in the field of Human-Human Interaction studies \cite{kim2022jointengagementclassificationusing} 
and even more in HRI. For example, 
\cite{Poggi2022,Sorrentino24} defined engagement as \textit{“the value that a participant in an interaction attributes to the goal of being together [...] and of continuing the interaction"}.

This value can be measured as a binary state \cite{Ben_Youssef_2019, Lu2024Complex,Saleh2021,Leite2015}, on an ordinal \cite{kim2022jointengagementclassificationusing, RudovicActiveLearning2019} or continuous scale 
\cite{Del_Duchetto_2020}, or with event-based annotations with multiple classes \cite{benyoussef:hal-02943475}. 

To infer these responses, the methods explore a wide variety of inputs from physiological data \cite{RudovicCultureNet,Rudovic2018Autism}, audio \cite{Ben_Youssef_2019}, 
or 2D RGB video \cite{ryoo2013first, mohsen2025mintrvaemulticuesintentionprediction}, to 3D LiDARS, Sonars or RGB-D \cite{salam:hal-02422969, Vaufreydaz_2016, Ben_Youssef_2019} signals. Processing is usually done with learning based methods from either CNN features \cite{Del_Duchetto_2020,Saleh2021}, or with intermediate representations such as speech-to-text transcriptions \cite{Lu2024Complex, Ben_Youssef_2019}, pose \cite{Vaufreydaz_2016, bian2025interactmejointegocentric, Bi2023Fuzzy,Xu2013Multiparty}, 2D trajectory and orientation \cite{Kato2015MayIHelpYou, Bohus2009} and, in many instances, gaze \cite{Liu2018PredictingEB, Xu2013Multiparty,Jung25EyeContact,Castellano2009,sidner2005explorationsengagementhumansrobots}.

\subsection{Interaction anticipation}

Often phrased as predicting the \textit{intention to interact}, actual \textit{intention} of interacting is not objectively measurable, as it is a non-observable mental state \cite{bian2025interactmejointegocentric} and is even subject to bias with self-assessment in real-time \cite{mohsen2025mintrvaemulticuesintentionprediction}. Most methods use \textit{a posteriori} observations of the interaction to predict whether a person will interact in the near future based on weak signals occurring during this future time interval.


When processing 2D image signals, human posture is a common feature in many studies. Adding hand landmarks is of little help \cite{bian2025interactmejointegocentric}, as is emotion analysis \cite{mohsen2025mintrvaemulticuesintentionprediction}. Regarding the use of 3D information, the benefit of detailed information such as a Kinect-based skeleton \cite{pard_thompson} compared to a lower dimensional input that combines position, velocity, head and/or torso orientation \cite{pard_thompson,Abbate_2024} or even the simple distance information \cite{Abbate_2024} is debatable. 

Temporal modeling of this prediction can be achieved using machine learning approaches such as random forests \cite{pard_thompson,Abbate_2024} or deep learning approaches such as GRU \cite{pard_thompson}, LSTM \cite{Abbate_2024}, bidirectional LSTM \cite{bian2025interactmejointegocentric}, or lightweight Transformers \cite{mohsen2025mintrvaemulticuesintentionprediction} but also plain MLP \cite{Abbate_2024}. The spatial dependency of the joints can be modeled using GCN \cite{kipf2017semisupervisedclassificationgraphconvolutional,bian2025interactmejointegocentric}. 

For evaluation, this task is typically framed as a binary classification problem within a variable temporal window \cite{pard_thompson,Abbate_2024,mohsen2025mintrvaemulticuesintentionprediction}, evaluated using F1 \cite{pard_thompson,bian2025interactmejointegocentric}, Macro-F1 \cite{mohsen2025mintrvaemulticuesintentionprediction} or AUC / AUROC \cite{mohsen2025mintrvaemulticuesintentionprediction, Abbate_2024}. \cite{bian2025interactmejointegocentric} analyses the effect of the window size and reports only marginal improvement of their \textit{intent to interact} classification, but a more significant one for the auxiliary task of \textit{action classification}. \cite{Abbate_2024} also reports the \textit{advance detection time}: the average lead time before an interaction is detected and analyzed performance across distance bins to reduce bias from proximity cues.

\subsection{Related datasets}

Databases relevant to egocentric HRI anticipation are presented in Table~\ref{tab:comparison}. The interacting ego recording platform is always referred to as \textit{robot} but can be very diverse : coffee machine, quad wheeled robot, service robot \cite{Abbate_2024}, moving trashcan \cite{bu2024ssuphrisocialsignalingurban}, teddy bear \cite{ryoo2013first}, 
robotic arm \cite{mohsen2025mintrvaemulticuesintentionprediction}, anthropomorphic robot (Pepper) \cite{benyoussef:hal-02943475}. In our case, we use a custom-made service robot. Nevertheless, all of them are equipped with at a least one visible camera. Among all these datasets, some characteristics are crucial for perception techniques in the context of social robotics.

\textbf{In the wild data collection (ITW)} means that passersby can spontaneously interact with the robot or not \cite{pard_thompson,benyoussef:hal-02943475,Abbate_2024,BrscicKanda2013, Kato2015MayIHelpYou}, opposed to predefined interactions played out by actors or structured interactions in controlled experiments, which yield a very limited set of behaviors \cite{Vaufreydaz_2016, ryoo2013first,ryoo2015prediction, mohsen2025mintrvaemulticuesintentionprediction}. It should be noted that in free interactions, new users often display transient behaviors during a phase of curiosity, which is not representative of regular interactions and fades over time, making long-term data collection valuable.

\textbf{Dataset scale} is a particularly important criterion for deep learning. 
This comes at the cost of
annotation time, where automatic annotation based on criteria such as speed, orientation, and proximity to the robot \cite{pard_thompson, Hall1966HiddenDimension, Abbate_2024, benyoussef:hal-02943475} allows for the collection of larger volumes of labeled data, as opposed to manually annotated \cite{Del_Duchetto_2020,ryoo2015prediction,bu2024ssuphrisocialsignalingurban} or controlled-environments \cite{Vaufreydaz_2016,mohsen2025mintrvaemulticuesintentionprediction} databases.

\textbf{Environment diversity} is crucial for robust learning based classifiers regardless of the context, since geometric based features (2D or 3D position of human body joints or trajectories of those) exhibit specific patterns that are highly dependent on the geometric setup of the environment, as illustrated in Figure \ref{fig:heatmaps_all}. 


Among \textit{in the wild} datasets, none of them provide simultaneously enough various scenes and sufficient interaction samples. 
\cite{pard_thompson} proposed two environments, but they are very similar and their field of view is too restrictive.
Instead, we propose a large, varied, curated, and in-the-wild 360 field of view dataset, named HUI360, supplemented by the homogeneous annotation of the videos of SSUP-HRI \cite{bu2024ssuphrisocialsignalingurban} that enables direct transfer evaluation of approaches.

\section{DATASET}
\label{sec:methods}
\subsection{Overview}





The dataset is organized in 68 recordings (\textbf{R}) that range from 15 minutes to 4 hours. Each recording consists of 3 to 134 episodes (\textbf{E}). An episode starts 10 second before a person is detected in front of the camera and ends after 10 seconds without detection in front of the camera. In a given environment the robot may have been placed differently from one recording to another when they were happening on different days. We distinguish these slight variations of the robot position and orientation by reporting the number of different setups \textbf{S} per environment. 
On average,  we recorded 478 tracks (maximum 839) per environment and 144 tracks (maximum 485) per unique setup. 

\begin{table*}[ht]
\centering
\caption{Description of the HUI360 and SSUP-A dataset. \textbf{S}: Setups, \textbf{R}: Recordings, \textbf{E}: Episodes, \textbf{(b.i.)} : before interacting, \textbf{\#Frames} refers to the number of frames where at least one person is present. In HUI360 out of 71.3 hours of recording we extracted 11.3h of episodes at 15fps (612K frames). AstorPlace and AlbeeSquare refer to the two field sessions of SSUP-HRI \cite{bu2024ssuphrisocialsignalingurban} each of them spanning several days with a mobile (\textit{M}) robot. We create train and validation set for both dataset and refer to them as \textbf{HUI360 Train} \colorbox{Emerald}{1}-\colorbox{Emerald}{7}, \textbf{SSUP-A Train} \colorbox{Emerald}{10}, \textbf{HUI360 Test} \colorbox{OrangeRed}{8}\colorbox{OrangeRed}{9} and \textbf{SSUP-A Test} \colorbox{OrangeRed}{11}.}
\fontsize{8}{10}\selectfont
\begin{tabular}{lcccccccccc}
\toprule
\multirow{3}{*}{\textbf{\makecell[c]{Environment}}} & \multirow{3}{*}{\textbf{S}} & \multirow{3}{*}{\textbf{R}} & \multirow{3}{*}{\textbf{E}} & \multirow{3}{*}{\makecell[c]{\textbf{Total} \\ \textbf{duration} \\ \textbf{(hours)}}} & \multirow{3}{*}{\textbf{\#Frames}} & \multirow{3}{*}{\makecell[c]{\textbf{Avg. tracks} \\ \textbf{per frame} \\ \textbf{(max)}}} & \multicolumn{2}{c}{\multirow{1}{*}{\textbf{\#Tracks}}} & \multicolumn{2}{c}{\multirow{1}{*}{\textbf{Tracks duration (s)}}} \\
\cmidrule(lr){8-9}\cmidrule(lr){10-11}
{\rule{0pt}{3ex}} & {} & {} & {} & {} & {} & {} & \textbf{All} & \textbf{Interacting} & \textbf{All} & \textbf{Interacting (b.i.)} \\
\midrule
\textbf{\colorbox{Emerald}{1} Bulle12X} & 1 & 3 & 230 & 7,4 & 43 373 & 1,2 (5) & 331 & 39 (12\%) & 10,4 $\pm$ 6,4 & 4,5 $\pm$ 2,3 \\
\textbf{\colorbox{Emerald}{2} Cafeteria} & 3 & 7 & 283 & 8.6 & 113 662 & 2,1 (12) & 993 & 74 (7\%) & 16,2 $\pm$ 19,7 & 13,3 $\pm$ 13,1 \\
\textbf{\colorbox{Emerald}{3} CoffeeB} & 3 & 8 & 263 & 8,8 & 119 886 & 1,6 (9) & 647 & 48 (7\%) & 19,6 $\pm$ 19,4 & 13,1 $\pm$ 12,5 \\
\textbf{\colorbox{Emerald}{4} ECBack} & 8 & 17 & 391 & 16,7 & 106 149 & 1,7 (8) & 763 & 80 (10\%) & 16,2 $\pm$ 28,1 & 10,4 $\pm$ 10 \\
\textbf{\colorbox{Emerald}{5} MainEntrance} & 4 & 6 & 160 & 5,1 & 57 897 & 2,0 (9) & 493 & 43 (9\%) & 15,4 $\pm$ 25 & 11 $\pm$ 7,7 \\
\textbf{\colorbox{Emerald}{6} MainHallway} & 1 & 1 & 29 & 0.9 & 8 118 & 1,3 (7) & 50 & 2 (4\%) & 14,3 $\pm$ 6,7 & 7,6 $\pm$ 5,6 \\
\textbf{\colorbox{Emerald}{7} Room005} & 1 & 1 & 10 & 1,0 & 3 145 & 1,1 (2) & 11 & 3 (27\%) & 20,7 $\pm$ 10 & 4,2 $\pm$ 1,5 \\
\textbf{\colorbox{OrangeRed}{8} ECFace} & 4 & 6 & 94 & 4,2 & 23 242 & 1,6 (8) & 183 & 21 (11\%) & 13,9 $\pm$ 16,3 & 6 $\pm$ 3,8 \\
\textbf{\colorbox{OrangeRed}{9} Room104} & 5 & 19 & 477 & 18,5 & 136 906 & 1,4 (8) & 839 & 65 (8\%) & 15,5 $\pm$ 20 & 6,3 $\pm$ 5,6 \\
\midrule
\textbf{Total (HUI360)} & 30 & 68 & 1 937 & 71,3 & 612 378 & 1,7 (12) & 4 310 & 375 (9\%) & 15,9 $\pm$ 21,3 & 9,8 $\pm$ 9,9 \\
\midrule
\textbf{\colorbox{Emerald}{10} AstorPlace} & \textit{M} & 16 & 118 & 9,8 & 519 521 & 5,1 (17) & 12 704 & 218 (2\%) & 13,9 $\pm$ 13,9 & 10,6 $\pm$ 13,4 \\
\textbf{\colorbox{OrangeRed}{11} AlbeeSquare} & \textit{M} & 12 & 159 & 12,1 & 629 917 & 5,8 (19) & 14 994 & 201 (1\%) & 16,3 $\pm$ 20,3 & 18,7 $\pm$ 23,1 \\
\midrule
\textbf{Total (SSUP-A)} & \textit{M} & 28 & 277 & 21,9 & 1 149 438 & 5,5 (19) & 27 698 & 419 (2\%) & 15,2 $\pm$ 17,7 & 14,5 $\pm$ 19,1 \\
\midrule
\textbf{Total} & 30 & 96 & 2 214 & 93,3 & 1 761 816 & 4,2 (19) & 32 008 & 794 (2\%) & 15,3 $\pm$ 18,2 & 12,3 $\pm$ 15,6 \\
\bottomrule
\end{tabular}
\label{tab:datasets_description}
\end{table*}

\subsection{Dataset Collection}
\subsubsection{HUI360: Diversity in-the-wild}

The data collection received the approbation of the Anonymous Institute Ethics Committee; we refer to the Supplementary Material for details regarding the compliance measures. Our data collection took place at 9 different places (referred to as \textbf{Environments}).
There were no instructions regarding the attitude to take in front of the robot, which was still, offering various objects to the participants (food, stickers, pins). The deployment was conducted in a total of 20 days across 3 months.

The dataset contains a wide variety of behaviors and appearances both for interacting and non interacting individuals and groups, as illustrated in Figure \ref{fig:groups} and Figure \ref{fig:indiv}.

With HUI360, we aim at fostering the development of methods for anticipating the interaction by modelling and understanding the human behavior beyond its geometric path with regard to a fixed distribution in a given setting. Such an achievement opens the door to deployment of systems capable of anticipating human behavior and behaving accordingly without waiting for retraining \cite{Abbate_2024}, thus enabling mobile applications such as the one of SSUP. To this end, we conduct our recordings in 9 different environments exhibiting different distributions of trajectories (cf. \ref{fig:heatmaps_all}), background\footnote{Empty background images of each environment are also openly released in HUI360.} and individual / group dynamics. 

\subsubsection{SSUP-A: Extension to other domains}

To push even further the diversity of environments, we annotated the SSUP-HRI dataset \cite{bu2024ssuphrisocialsignalingurban}, which was collected in the wild during 2 field sessions of 5 days each, one year apart, on public squares of New York using 2 mobile trash barrels equipped with 360 cameras and operated in a \textit{Wizard of Oz} fashion. The videos exhibit a full range of passerbys and users all around, with individual and group interactions. The different nature of behaviors, the shift in viewpoint, as well as the use of a mobile recording platform make it an interesting testbed for benchmarking the transferability of HRI anticipation methods to drastically different domains. We detected, tracked and annotated an average of 5.5 persons per frame, on a total of 1.15M frames.

For practical reasons, we split the SSUP dataset in 5 minutes episodes, meaning that some tracks are interrupted and restarted as a new one in the next episode.


\subsection{Pre-processing and automatic labelling}
\subsubsection{Detection and tracking}

We detected persons using YOLOv11x \cite{yolo11_ultralytics} and used SAM2.1-L \cite{ravi2024sam2} for tracking and segmentation across an episode, as both have shown very good results even in the presence of occlusions and with a moving camera. To avoid tracking from low quality or partial body detection, we performed a filtering after detection, based on valid visible pose keypoints.

\begin{figure}[htbp]
    \centering
    \includegraphics[width=0.48\textwidth]{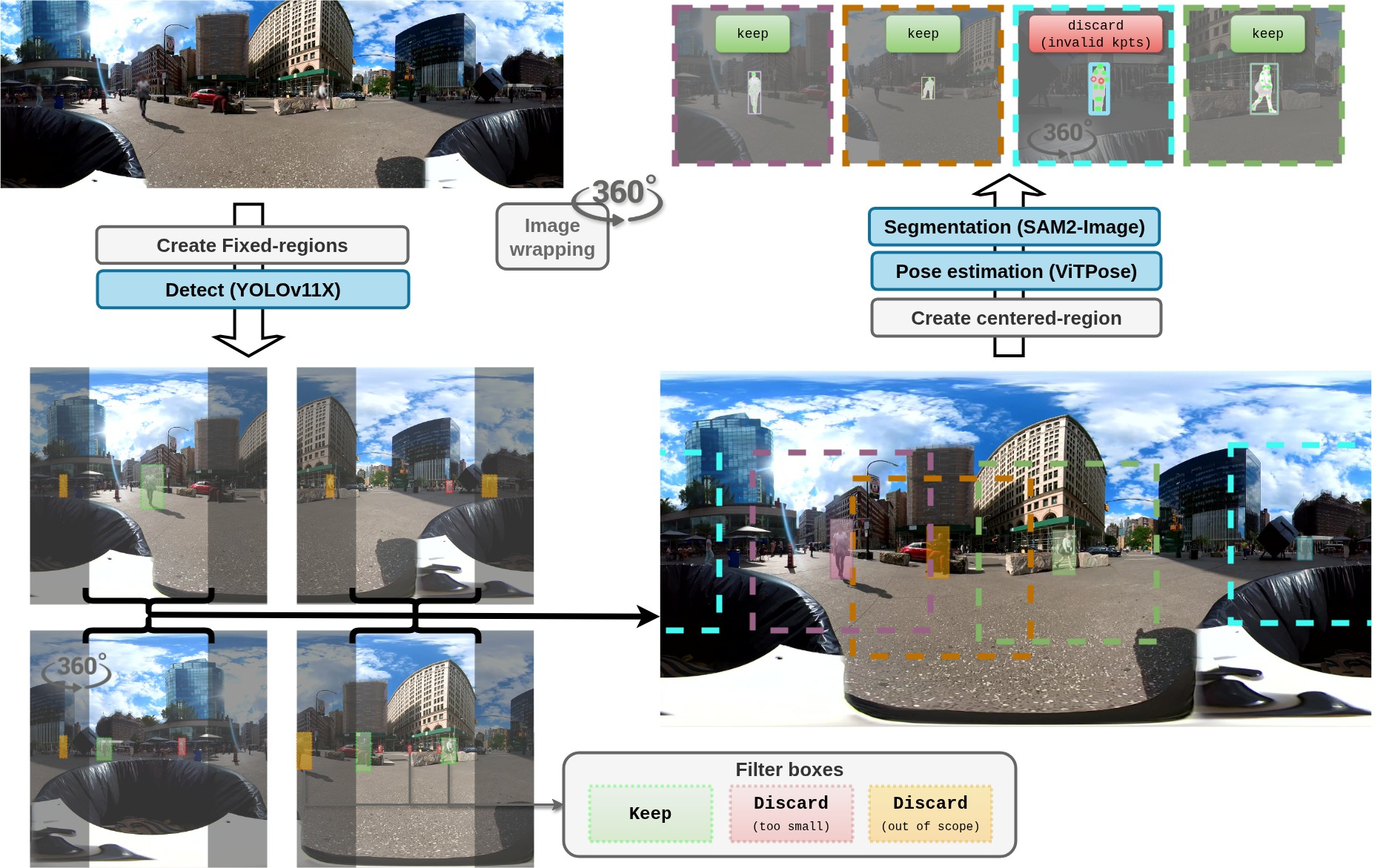}
    \caption{Detection and filtering process in equirectangular images: the process of detection uses 4 overlapping fixed regions of the image with wrapping. Filtering is done based on box size, mask size and number of valid keypoints.}
    \label{fig:detecting}
\end{figure}

In practice, we performed the steps of detection, filtering, segmentation and tracking using region-images (crops of the panoptic image) to mitigate the negative effects of directly using panoptic images on models that have not been trained with them. The full process is briefly illustrated in Figure \ref{fig:detecting} (detection and filtering) in Figure \ref{fig:tracking} (tracking and segmentation) and is detailed in the supplementary materials (Algorithm \ref{alg:panoptic}).

\begin{figure*}[htbp]
    \centering
    \includegraphics[width=\textwidth]{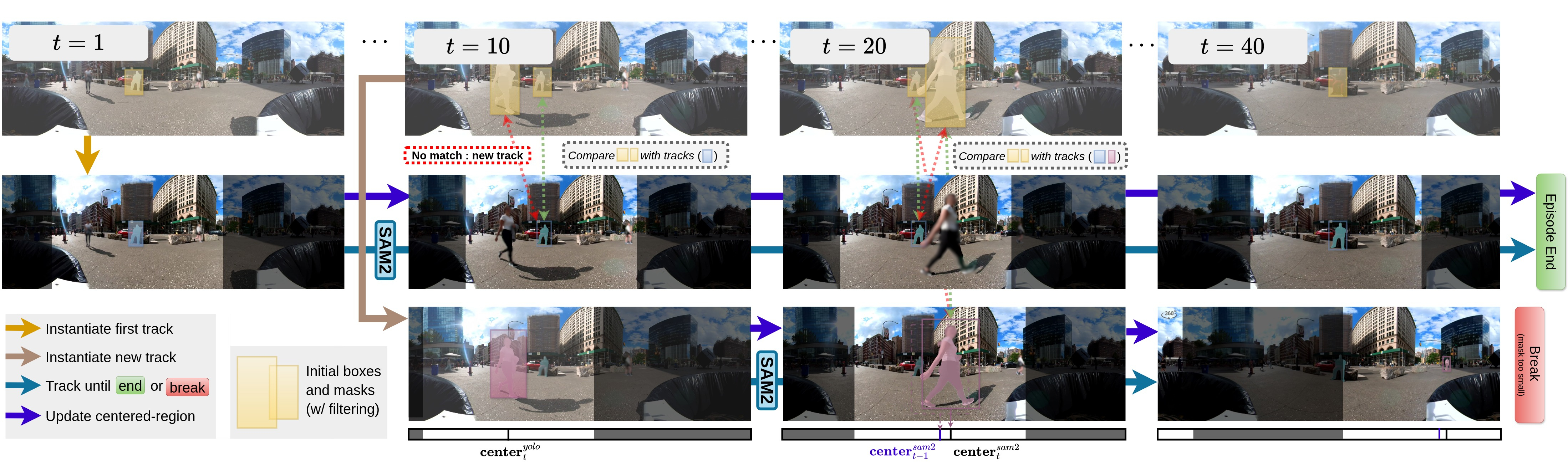}
    \caption{Tracking and segmentation pipeline in equirectangular images: using SAM2 new tracks are instantiated from the filtered detections and processed using a mobile region of the image until it breaks or reaches the end of the episode - then the next track is instantiated from the next non-overlapping detection.}
    \label{fig:tracking}
\end{figure*}

\subsubsection{Automatic detection of interactions}

We adopt a precise definition of an interaction where we only consider physical interaction with the platform. In practice, we define an interaction zone (the trashcans in SSUP-A and the plate of the robot in HUI360) and consider that someone is interacting when their segmentation mask intersects with this zone. Doing so is straightforward in HUI360, as the interaction zone occupies a fix place in the camera frame, but the mounting of the camera on the trashcans of SSUP makes that they move and appear differently in the camera frame; in consequence, we use a different intersection mask for each frame obtained with SAM2 with a manual prompt of the initial position and refined by computing its Convex Hull, such that, when someone throws something or puts their hand in the trashcan, we can still compute the intersection, see this process in the Supplementary Materials (Figure \ref{fig:convexhull}).

Our method can then be used for any fixed or moving interaction zone as long as it is a convex shape. 

    

\begin{figure}[htbp]
    \centering
    
    \begin{subfigure} 
        \centering
        \includegraphics[width=\linewidth]{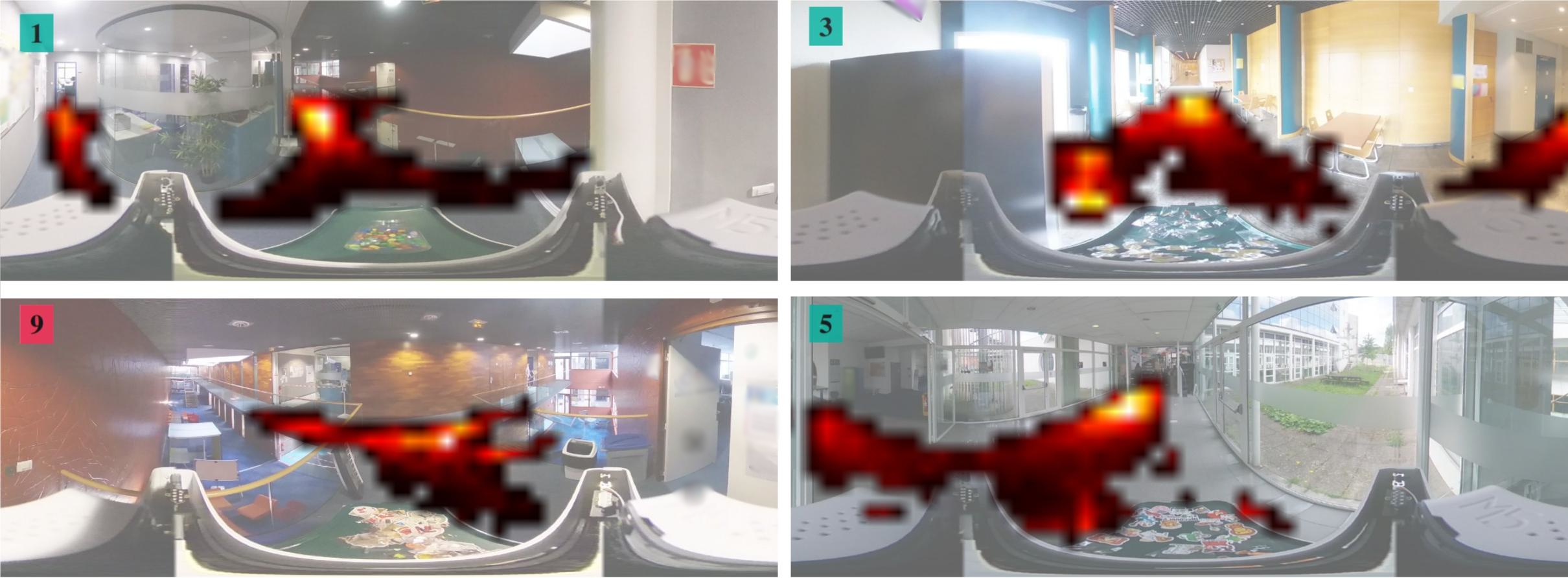}
    \end{subfigure}
    
    
    \begin{subfigure}
        \centering
        \includegraphics[width=\linewidth]{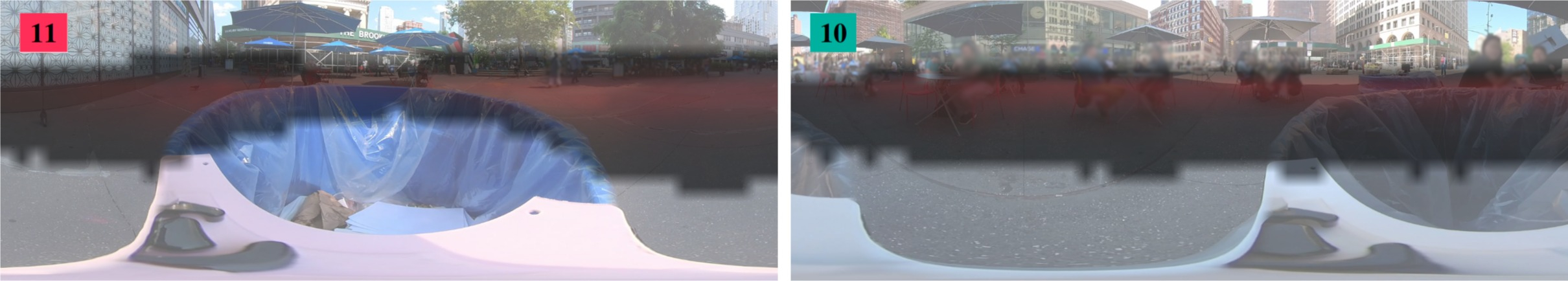}
    \end{subfigure}

    \caption{Heatmaps of the ankles position of all tracks in  HUI360 (top two rows) and SSUP-A (bottom row). Note that data in SSUP-A are recorded with moving robots, the background images only serve as illustrations and are not geometric references.}
    \label{fig:heatmaps_all}
\end{figure}

\subsubsection{Automatic extraction of pose}

For each detection of each track we extracted the pose of the person using two methods : 
\begin{itemize}
    \item ViTPose-B \cite{xu2022vitpose} that achieves near state of the art performances on benchmarks while being very efficient. This method yields 17 keypoints with the COCO pose format.
    \item Sapiens-0.6B-Pose-308 \cite{khirodkar2024sapiensfoundationhumanvision} a finetuned version of Sapiens-0.6B, a foundation model for various downstream human vision tasks. In this version the model is finetuned on 1M manually annotated images to predict 308 keypoints including 242 facial keypoints. 
\end{itemize}

Two methods were used because they appear to be complementary. Indeed, ViTPose operates reliably and robustly at reasonable working resolutions ($256\times 192$), whereas Sapiens works at much higher resolutions ($1024 \times 768$) but provides a very precise estimation with keypoints on the face and hands.
In our dataset 26\% of the boxes have and height $h < 256$, 73\% have $h \in [256,1024]$ and 0.4\% have $h > 1024$ (after $\times$1.25 expansion as used for those methods). Meaning that most of the persons images are downscaled for ViTPose and upscaled for Sapiens, ultimately 73\% of ViTPose and 62\% of Sapiens keypoints (64\% for facial keypoints, and 55\% for the rest) are valid (i.e. with a score $\mathsf{c} > 0.5$). 


\subsection{Manual curation}

To ensure the quality of the data we share, we proceeded to manual curation using specifically designed user interfaces. Our curation consisted in \textbf{(1)} flagging full episodes as invalid for reason such as: multiple abnormal behaviors attributed to the novelty of the robot's presence, presence of the operator for technical task (launching/stopping recording, maintaining the robot), multiple and non recoverable tracking issues, removal request in compliance with our ethics protocol and GDPR, \textbf{(2)} corrections on tracking : removing, splitting or associating tracks, \textbf{(3)} corrections on automatic labeling (especially in SSUP-A) : correcting false negatives (interaction without masks intersection such as throwing or dropping an object without direct contact) and false positives (mostly induced by an excessive interaction zone mask when it appears as not convex because of the moving 360 camera view point). We refer to the Supplementary Materials (Figure \ref{fig:novelty}) for examples of such issues that were flagged or corrected. Overall 15\% of episodes were flagged and not exported.


\subsection{Problem statement}
\label{ssec:pb_statement}

Since we cannot observe the actual \textit{intent of a person to interact} at a given time, we formulate our task as \textit{interaction anticipation} for more clarity. 
In this work, we deliberately focus on a restricted class of interactions defined as \textit{physical interactions} with the robot platform (e.g., picking, placing, or throwing objects). 
This choice is motivated by two key properties: (1) such interactions can be automatically detected and labeled at scale using geometric criteria (mask intersection with a predefined interaction zone), enabling the construction of large datasets; and (2) they provide an objective and reproducible definition of interaction onset, avoiding ambiguity and inter-annotator variability inherent to more subjective cues such as gaze, hesitation, or verbal engagement.

Formally, we consider a set of $N$ inputs $\mathcal{X} = \{X_1,...,X_N\}$ and associated labels $\mathcal{Y} = \{y_1,...,y_N\}$ with $y_i\in\{0,1\}$ where a positive label $y_i=1$ corresponds to a sequence from which a future interaction should be anticipated. Each input $X_i$ is a set of features observed within $T$ frames, referred to as \textit{observation window}, such that $X_i = [x_{i,1},...,x_{i,T}] \in \mathbb{R}^{T\times D}$, $T = \ceil{\frac{\mathbf{T}^{org}}{\mathbf{s}}}$ with $s$ a subsampling factor, and $D$ the dimension of features.

Formally we define $\mathbf{T_0}$ the beginning time of an interaction and $\mathbf{T_{POS}}$ the positive label cutoff, that is the time (in number of frames) before the interaction for which any observation windows including this time is considered as positive. Additionally, $\mathbf{T_{CUT}}$ is defined as a second cutoff such that all segments that end up being to close to the actual interaction are discarded. With these two cutoff thresholds multiple observation windows can be defined for one interaction event, which could be very useful at training time as a data augmentation. Nevertheless, at training time, we choose to draw only one observation windows positioned at $\mathbf{T_{ADV}}$ corresponding to $\mathbf{T_{ADV}} = \mathbf{T_{POS}} = \mathbf{T_{CUT} + 1}$.
With this convention, an input sample is assigned a positive label if the last frame of its observation window occurs at a time $\mathbf{T_{ADV}}$  prior to the interaction onset time $\mathbf{T_{POS}}$. Any temporal window whose endpoint precedes $\mathbf{T_0} - \mathbf{T_{ADV}}$ is labeled as negative. Temporal windows whose endpoints fall within the interval $[\mathbf{T_0} - \mathbf{T_{ADV}}, \mathbf{T_0}]$ are excluded from evaluation, as they are assumed to lie outside the anticipatory regime.

For negative samples, the same temporal rationale applies; however, candidate $\mathbf{T_0}$ instances must be selected. Rather than sampling them randomly, we define $\mathbf{T_0}$ as the moment at which the tracked subject appears largest in the field of view. Empirically, this serves as a reliable proxy for estimating the subject-to-robot distance (more details in the Supplementary Material). This way we don't use obvious negatives that are \textit{moving away} from the robot and mitigate imbalance between overwhelming non interacting tracks and interacting ones.

The HUI360 and SSUP-A dataset are partitioned into training and testing splits as indicated in Table \ref{tab:datasets_description}. Two out of the nine environments for HUI360 dataset and the second half of SSUP-A dataset are kept for evaluation. 
This dataset come along with a series of evaluation baselines. We mainly focus on 3 type of evaluations: 
\begin{itemize}
  \item \textbf{Transfer capabilities}: first, within each dataset, by training and evaluating on disjoint environments while keeping the same robot platform; then, through direct cross-dataset transfer, thereby enabling analysis of the impact of robot variation (e.g., size, mobility characteristics, functional role) on anticipation performance. This evaluation is done with $\mathbf{T_{ADV}} = 15$.
  \item \textbf{Forecasting capabilities}: by varying $\mathbf{T_{ADV}} \in [5, 10, 15, 20, 30]$ corresponding respectively to $[0.33, 0.66, 1, 1.33, 2]$ second anticipation horizons, this evaluation give information of how far a method can anticipate an upcoming interaction.
  \item \textbf{Input frequency robustness}: under intentionally temporally downsampled input conditions. Using $\mathbf{s}$, performances with an adapted $\mathbf{T_{ADV}} = 1$sec with input framerate equal respectively to $[15, 5, 3, 1]$Hz provide an important information on the behavior of the anticipation system under constrained resource conditions. This feature is particularly an important for robotic applications, where on-board computational power is inherently limited.
\end{itemize}







\section{EXPERIMENTS}
\label{sec:expes}

\subsection{Evaluated baselines}

We implemented three straightforward baselines inspired by architectures commonly used in recent works \cite{Abbate_2024, pard_thompson}: a Random Forest classifier (RF), a simple Multi-Layer Perceptron (MLP) and a Long Short-Term Memory RNN (LSTM). The first two models do not explicitly capture temporal dependencies, but they can leverage an arbitrary number of features aggregated over any reasonable fixed observation window while the LSTM provides a sequential model capable of encoding motion over variable windows and remains lightweight under higher framerate.
For the input representation, we selected a combination of the segmented person’s area, used as a proxy for distance, its bounding box position, and the ViTPose body keypoints (see the ablation study in \ref{ssec:ablative_studies}). 


Regarding evaluation, we report AUC as a primary metric. While it has limitations under class imbalance (with a positive/negative ratio of 0.19 in HUI360 Test and 0.03 in SSUP-A Test), it provides a threshold-independent view of performance, which is suitable since the optimal operating point depends on the application.

In practice, threshold selection reflects a trade-off between false positives (which could lead to over-engaging users) and false negatives (missing interactions). We therefore favor a threshold-free metric that captures ranking ability.

For completeness, we also report F1-score and Average Precision (\ref{tab:transfer_results_apf1}) for the main cross-dataset evaluation (\ref{tab:transfer_results}). These confirm the trends observed with AUC, while making the impact of class imbalance more visible, especially on SSUP-A.

\subsection{Implementation details}

\textbf{Classifiers} Our simple MLP classifier used 2 hidden layers with ReLU activation (sizes $[32,256]$). The LSTM comprised 3 layers with a hidden size of $128$ and a linear classifier at the end. At $T=30$ the MLP comprises between 9k and 0.9M and for any $T$ the LSTM between 350k and 0.85M trainable parameters, depending on the chosen $D$. Both were trained during 25 epochs using weighted Binary Cross Entropy (BCE) loss, a fixed learning rate of $0.001$, we did not apply dropout. 


The RF classifier used a limited depth $d_{RF}=5$ and an important number of trees $N_{RF}=500$, we also applied class weight balancing to account for the highly imbalanced label distribution during the construction of the trees.


\textbf{Normalization} Inputs are standardized using mean and standard deviation from the whole HUI360 dataset. Previous to standardization all keypoints are normalized with regard to their associated bounding box.

\textbf{Sampling from tracks}
In our HUI360 and SSUP-A we are given $N_{raw}$ tracks: $\mathcal{T} = \{\mathcal{T}_1,...,\mathcal{T}_{N_{raw}}\}$ such that  $\mathcal{T}_i = [x_{i,1},...,x_{i,T_i}] \in \mathbb{R}^{T_i\times D}$ and \textit{per frame} automatic label indicating if the track is \textit{currently} interacting $\{[a_{i,1},...,a_{i,T_i}]\}_{i=1}^{i=N_{raw}}$ with $T_i$ the total length of the track. %

\begin{figure}
    \centering
    \includegraphics[width=\linewidth]{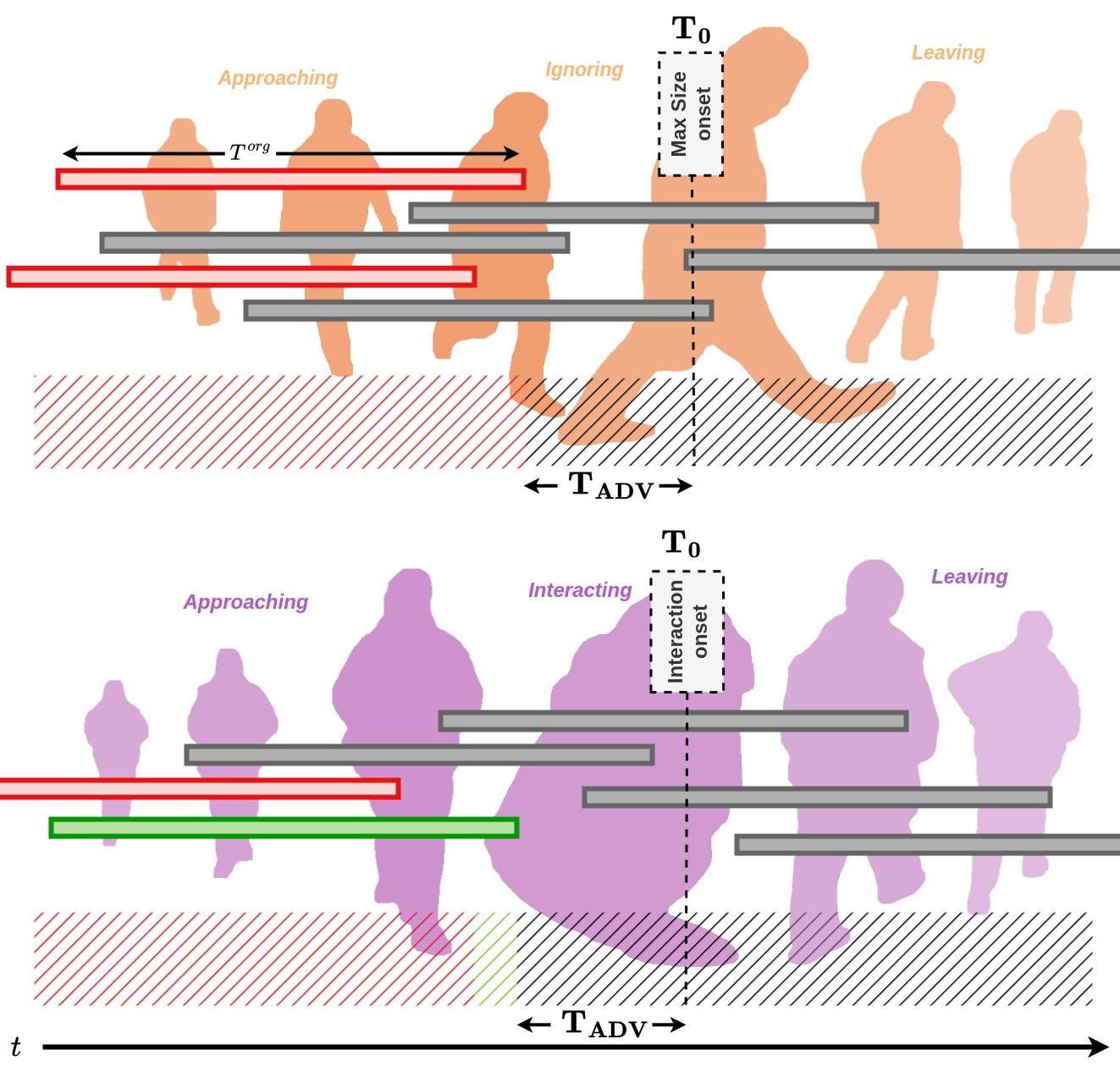}
    \caption{Temporal rationale of input labelling and sampling. On top an non interacting track. Segments in \colorbox{green}{green} represents possible positive inputs ($y_i = 1$), segments in \colorbox{red}{red} are possible negative inputs ($y_i=0$) and segments in \colorbox{gray}{gray} are discarded during filtering.}
    \label{fig:tracks_logic}
\end{figure}

Creating segment to sample from these data to generate $\mathcal{X}$ and $\mathcal{Y}$ requires some heuristics that we explicit in the supplementary materials (Algorithm \ref{alg:track_sampling}) and illustrate briefly in Figure \ref{fig:tracks_logic}

\subsection{Results}\label{sec_results}
\subsubsection{Main baselines}

We provide 3 main baselines on \textit{transfer capabilities} (in-dataset : different scenes and layouts among the same dataset and cross-dataset : cross environment, sensor position, embodiment and human actions), \textit{forecasting capabilities} and \textit{input frequency robustness}.

Unless stated otherwise, we used the feature set $\mathcal{D}_3$ for all experiments and $\mathbf{T_{ADV}}=15$.
\vspace{8pt}

\textbf{Transfer capabilities} We assess the generalization performance of the proposed methods at two distinct levels. The first level focuses on environmental variation, evaluated on the respective test sets of the two datasets, each constructed to be disjoint from its corresponding training environments. The second level involves cross-domain evaluation, where—beyond changes in environment and without any additional fine-tuning—the robotic platform itself differs, introducing further domain shifts.

As reported in Table~\ref{tab:transfer_results}, the LSTM model outperforms the MLP and Random Forest in all scenarios. As expected, a change in the robotic platform results in a noticeable drop in performance under cross-dataset evaluation. 

These findings establish an initial baseline emphasizing generalization for robots operating in unseen, in-the-wild environments.

\begin{table}[ht]
\centering
\caption{AUC of MLP, RF and LSTM. Each model is trained on its native dataset, and subsequently evaluated in a cross-domain transfer scenario.}
\fontsize{8}{10}\selectfont
\begin{tabular}{ccccccc}
\toprule
{} & \multicolumn{3}{c}{HUI360 Test} & \multicolumn{3}{c}{SSUP-A Test} \\
\cmidrule(lr){2-4}\cmidrule(lr){5-7}
{} & \textbf{RF} & \textbf{MLP} & \textbf{LSTM} & \textbf{RF} & \textbf{MLP} & \textbf{LSTM} \\
\midrule
HUI360 Train & 0.81 & 0.86 & \textbf{0.91} & 0.83 & 0.79 & \textbf{0.84} \\
SSUP-A Train & 0.76 & 0.81 & \textbf{0.84} & 0.86 & 0.85 & \textbf{0.88 }\\
\bottomrule
\end{tabular}
\label{tab:transfer_results}
\end{table}


\begin{table}[ht]
\centering
\caption{Average Precision and F1-score (at 0.5) of MLP, RF and LSTM in in-dataset and cross-dataset evaluation}
\fontsize{8}{10}\selectfont
\begin{tabular}{ccccccc}
\toprule
{} & \multicolumn{3}{c}{HUI360 Test} & \multicolumn{3}{c}{SSUP-A Test} \\
\cmidrule(lr){2-4}\cmidrule(lr){5-7}
{} & \textbf{RF} & \textbf{MLP} & \textbf{LSTM} & \textbf{RF} & \textbf{MLP} & \textbf{LSTM} \\
\midrule
\multicolumn{7}{c}{AP} \\
\midrule
HUI360 Train & 0.48 & 0.53 & \textbf{0.60} & 0.14 & 0.12 & \textbf{0.16} \\
SSUP-A Train & 0.33 & 0.49 & \textbf{0.50} & 0.16 & 0.21 & \textbf{0.24} \\
\midrule
\multicolumn{7}{c}{F1-Score} \\
\midrule
HUI360 Train & 0.45 & 0.57 & \textbf{0.61} & 0.06 & 0.20 & \textbf{0.23} \\
SSUP-A Train & 0.39 & 0.44 & \textbf{0.50} & 0.23 & 0.24 & \textbf{0.26} \\
\bottomrule
\end{tabular}
\label{tab:transfer_results_apf1}
\end{table}

\textbf{Forecasting capabilities} On Table~\ref{tab:ablation_tADV}, we provide the baseline of our proposed models. Unsurprisingly, the closer the observation window is to the onset of the interaction, the easier it becomes to correctly classify the anticipation. Moreover, the decrease in performance follows a fairly regular trend. It can also be observed that the MLP and LSTM outperforms the Random Forest and exhibit a more gradual degradation in performance as $\mathbf{T_{ADV}}$ increases.

Classifiers and trained and evaluated with the same $\mathbf{T_{ADV}} \in \{5,10,15,20,25,30\}$, corresponding to a forecasting of 0.33, 0.66, 1.0, 1.33, 1.66 or 2.0 seconds respectively. For consistency all of them are only trained and evaluated on set of tracks from HUI360 Train and evaluated on a set of tracks from HUI360 Test that are long enough to allow sampling with $\mathbf{T_{ADV}}$ up to $30$ (for this reason results at $\mathbf{T_{ADV}}=15$ i.e. the default modality, differ from the other baselines.)


\begin{table}[ht]
\centering
\caption{AUC of RF and MLP classifier at different advance detection thresholds $\mathbf{T_{ADV}}$ (5 to 30 frames, or 0.33 to 2.0 seconds).}
\fontsize{8}{10}\selectfont
\begin{tabular}{ccccccc}
\toprule
$\mathbf{T_{ADV}}$ & 5 & 10 & 15 & 20 & 25 & 30 \\
\midrule
\textbf{RF} & 0.9 & 0.84 & 0.81 & 0.78 & 0.73 & 0.71 \\
\textbf{MLP} & 0.96 & 0.92 & 0.89 & 0.86 & 0.84 & 0.76 \\
\textbf{LSTM} & 0.97 & 0.94 & 0.89 & 0.87 & 0.84 & 0.77 \\
\bottomrule
\end{tabular}
\label{tab:ablation_tADV}
\end{table}

\textbf{Input frequency robustness} 
As anticipated, performance degrades as the input framerate decreases (cf. \ref{tab:ablation_subsampling}), with a pronounced collapse when the rate drops to a critical value of one frame per second. The results presented here serve as an additional baseline, highlighting the sensitivity of anticipation systems to temporal resolution under computational constraints.


\begin{table}[ht]
\centering
\caption{AUC of RF and MLP with different subsampling rate $\mathbf{s}$. For all baselines we keep $\mathbf{T^{org}} = 30$ (2 sec).}
\fontsize{8}{10}\selectfont
\begin{tabular}{ccccc}
\toprule
$\mathbf{s}$ / Freq. & 1 / 15Hz & 3 / 5Hz  & 5 / 3Hz  & 15 / 1Hz \\
\midrule
\textbf{RF} & 0.81 & 0.80 & 0.80 & 0.73 \\
\textbf{MLP} & 0.87 & 0.87 & 0.85 & 0.75 \\
\textbf{LSTM} & 0.91 & 0.87 & 0.85 & 0.78 \\
\bottomrule
\end{tabular}
\label{tab:ablation_subsampling}
\end{table}

\subsubsection{Ablation studies}
\label{ssec:ablative_studies}

We conducted an ablation study on various sets of features that we denote $\mathcal{D}_1$ to $\mathcal{D}_7$ and first need to be defined.


Since we don't have a distance information, we rely on the segmentation mask size to give us a proxy of it: 

\begin{itemize}[leftmargin=+.3in]
    \item $\mathcal{D}_1 = [m] \in \mathbb{R}$
\end{itemize}
Then we consider adding incrementally the box coordinates $B = [x^{min},y^{min},x^{max},y^{max}]$ normalized in the image frame, the 17 keypoints from ViTPose, the 242 facial keypoints and the rest of the body keypoints from Sapiens-308.

\begin{itemize}[leftmargin=+.3in]
    \itemsep0.2em 
    \item $\mathcal{D}_2 = [m,B] \in \mathbb{R}^5$
    \item $\mathcal{D}_3 = [\mathcal{D}_2, K^{vitpose}]\in \mathbb{R}^{56}$
    \item $\mathcal{D}_4 = [\mathcal{D}_3,K^{sapiens}_{\mathrm{face}}]\in \mathbb{R}^{782}$
    \item $\mathcal{D}_5 = [\mathcal{D}_3,K^{sapiens}]\in \mathbb{R}^{980}$
\end{itemize}
We added two other set of features : a lower dimensional handcrafted set of keypoints, with the shoulders, ears and eyes positions, considering that those may be informative on the orientation of the users with regard to the camera, and a set that includes ViTPose keypoints without the box position.

\begin{itemize}[leftmargin=+.3in]
    \itemsep0.2em 
    \item $\mathcal{D}_6 = [\mathcal{D}_2,K^{vitpose}_{\mathrm{should}},K^{vitpose}_{\mathrm{ears}},K^{vitpose}_{\mathrm{eyes}}] \in \mathbb{R}^{23}$
    \item $\mathcal{D}_7 = [m,K^{vitpose}] \in \mathbb{R}^{52}$
\end{itemize}
Note that since the keypoints are normalized in the detection box $\mathcal{D}_1$ and $\mathcal{D}_7$ do not retain information related to the position of the track in the frame.
The results presented in Table \ref{tab:ablation_features} suggest that the key points provided by Sapiens yield results similar to those of ViTPose for the reference databases tested, despite being much more detailed.

\begin{table}[ht]
\centering
\caption{Ablation study of feature sets for RF, MLP and LSTM classifiers. Best result for each classifier in \textbf{bold}.}
\fontsize{8}{10}\selectfont
\begin{tabular}{cccccccc}
\midrule
{} & $\mathcal{D}_1$ & $\mathcal{D}_2$ & $\mathcal{D}_3$ & $\mathcal{D}_4$ & $\mathcal{D}_5$ & $\mathcal{D}_6$ & $\mathcal{D}_7$ \\
\midrule
\textbf{RF} & 0.78 & 0.80 & 0.81 & 0.81 & \textbf{0.81} & 0.81 & 0.80 \\
\textbf{MLP} & 0.81 & 0.83 & 0.87 & 0.88 & 0.88 & 0.87 & \textbf{0.89} \\
\textbf{LSTM} & 0.79 & 0.81 & \textbf{0.91} & 0.88 & 0.90 & 0.90 & 0.89 \\
\bottomrule
\end{tabular}
\label{tab:ablation_features}
\end{table}

\section{CONCLUSION}
\label{sec:conclusion}












We introduced HUI360, the largest in-the-wild dataset for human–robot interaction anticipation, along with an automated pipeline for interaction detection. Recorded over multiple months and across diverse environments, it provides both curated annotations and access to raw 360° egocentric images, enabling research on new features, modalities, and applications.

Beyond its scale, HUI360 formalizes a consistent definition and evaluation protocol for interaction anticipation, offering a foundation for future comparative studies in computer vision and human–robot interaction.

We addressed the overlooked challenge of generalizing interaction anticipation methods to new environments by providing large scale annotations for another dataset and sharing baselines for both in-dataset cross-environment and cross-dataset zero-shot transfer scenarios. We further considered real-world deployment constraints by establishing results at reduced frame rates and across different forecasting horizons.

While the current baselines are architecturally basic, they open promising directions whether it is exploring richer temporal and spatial modeling, leveraging the specific characteristics of equirectangular imagery, or incorporating social and group dynamics. In this sense, HUI360 represents not only a significant step toward practical and generalizable human-robot interaction anticipation, but also an important resource for future research in socially aware robotics and anticipatory perception.

We acknowledge that the operational definition of interaction adopted in HUI360, based on physical contact with the platform, captures only a subset of the rich spectrum of human-robot interaction behaviors. In particular, it does not explicitly model pre-contact social signals such as gaze, hesitation, verbal engagement, or approach-and-stop behaviors, which may also convey interaction intent. 

This design choice reflects a trade-off between scalability and semantic richness. While objective and automatically measurable criteria enable large-scale and reproducible annotation, they also limit the range of interactions captured. We thus consider HUI360 as a foundation rather than a complete definition, and hope that the release of both the raw 360° data and the annotation pipeline will support future extensions toward more nuanced and socially grounded interaction labels.

\section{ACKNOWLEDGMENTS}

This work was partially supported by the European project euROBIN (Horizon Europe, GA. N. 101070596), the France 2030 program through the PEPR O2R projects AS3 (ANR-22-EXOD-007), and the Robotics Chair of the Cluster AI project ENACT (ENACT-ANR-23-IACL-0004). Experiments presented in this paper were carried out using the Grid'5000 testbed, supported by a scientific interest group hosted by Inria and including CNRS, RENATER and several Universities as well as other organizations (see \url{https://www.grid5000.fr}).




{\small
\bibliographystyle{ieee}
\bibliography{egbib}
}

\clearpage
\setcounter{page}{1}
\maketitlesupplementary


\section{HUI360 : Diversity in-the-wild}

\begin{figure*}[htbp]
    \centering
    \includegraphics[width=0.95\linewidth]{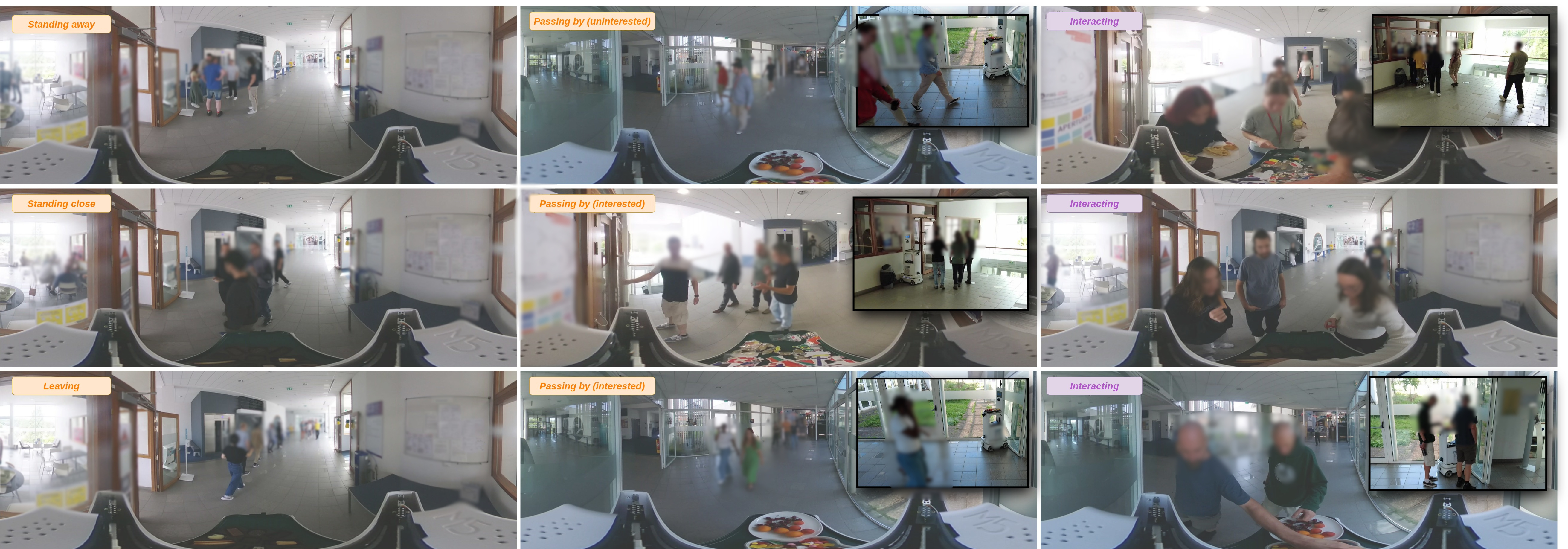}
    \caption{Diverse groups composition and behaviors. We show an external camera capturing the scene to make it easier for the reader to understand the group positions in front of the robot.}
    \label{fig:groups}
\end{figure*}

\begin{figure*}[htbp]
    \centering
    \includegraphics[width=0.95\linewidth]{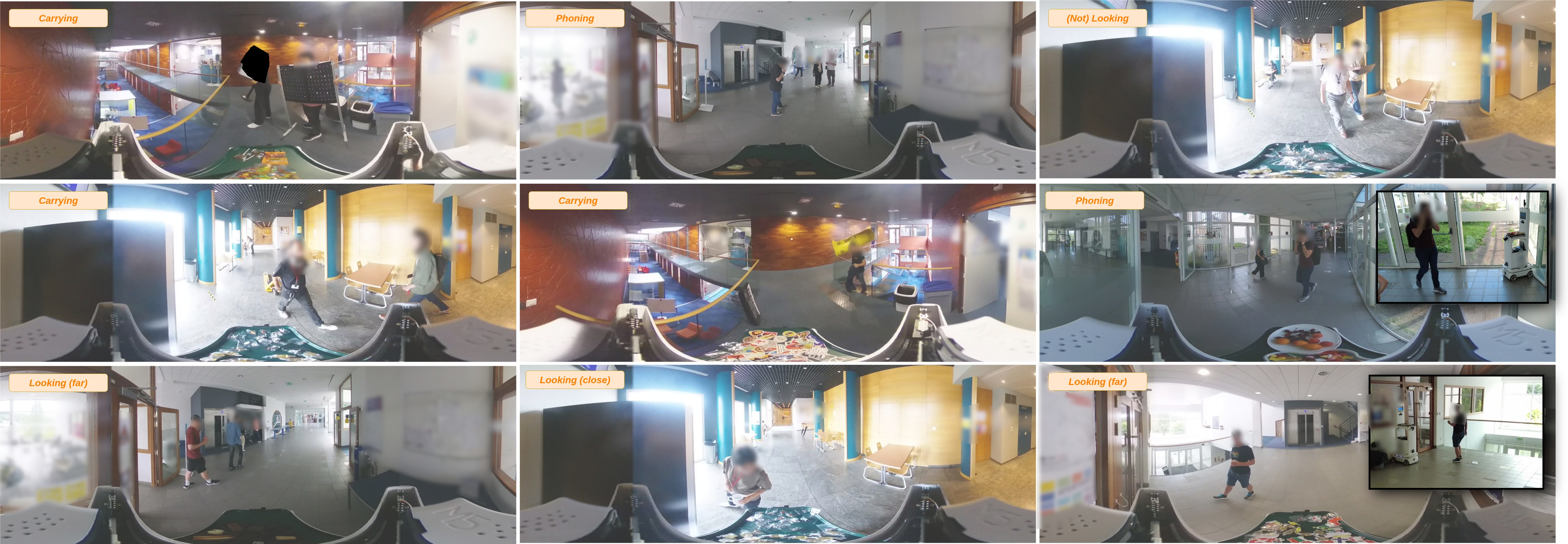}
    \caption{Diverse non-interacting individual behaviors. We show an external camera capturing the scene to make it easier for the reader to view the behavior of the passerby.}
    \label{fig:indiv}
\end{figure*}

\subsection{Data Collection: Ethics and Compliance}

The data collection process was approved by the Anonymous Institution Ethics Committee on June 17, 2025 (decision no. 465). To ensure natural interactions with the recording platform while complying with privacy and data protection regulations, all building employees were informed by email that recording activities would take place and were provided with information regarding their GDPR rights, including the right to withdraw their data. This information was also made accessible to passersby through poster signs displayed in close proximity to the robot.
Upon request, specific portions of the recordings were removed. Individuals who exercised their right to withdrawal were identified using ad-hoc software with automatic state-of-the-art facial recognition, followed by manual verification to ensure removal of those individuals from the dataset.

\subsection{Diversity}

HUI360 encompasses diverse behaviors for both interacting and non interacting tracks. Examples of such diverse behaviors for groups can be found for (\ref{fig:groups}) and individuals (\ref{fig:indiv}). 

\begin{figure}[H]
    \centering
    \includegraphics[width=0.95\linewidth]{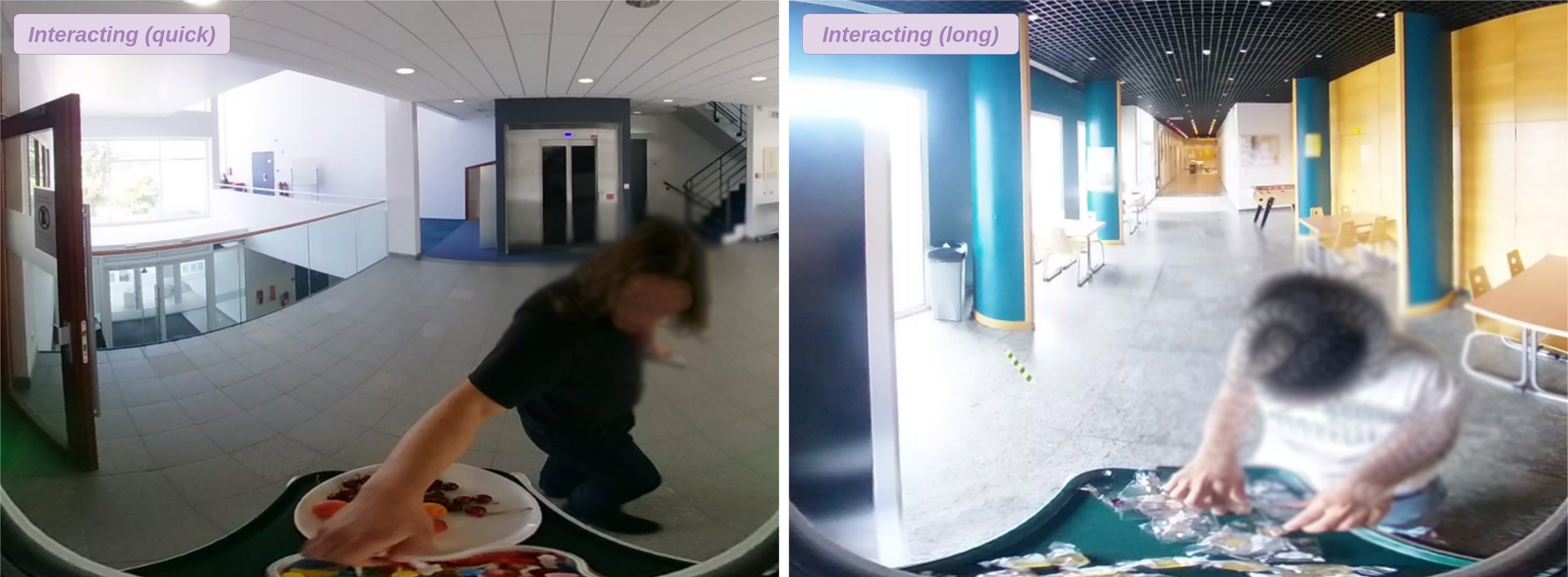}
    \caption{Diverse interaction durations}
    \label{fig:interaction_duration}
\end{figure}

Although our work focus on the \textit{initiation} of the interactions and not on the content and duration of the interactions themselves, it is noticeable that some interactions with the robot last longer (several seconds) than others (less than 5 frames that is 0.3sec) as illustrated in \ref{fig:interaction_duration}.

\section{Detection and tracking}

The detection and tracking pipeline is formally presented in \ref{alg:panoptic} to complement the illustration of Fig. 2 and Fig. 3, the pipeline consists of 4 steps: (1) detection and first basic filtering, using the detection box size and its associated confidence (\textbf{Step 1});  (2) initial person segmentation using SAM2.1-Image and additional filtering based on mask size in pixels (\textbf{Step 2}); (3) 2D Pose Estimation using ViTPose for further filtering, total number of valid keypoints and number of in-mask valid keypoints (\textbf{Step 3}); and finally (4) tracking and segmentation using SAM2.1-Video with breaking based on mask size in pixels or overlap with existing detections (\textbf{Step 4}).

\textbf{Step 1} and \textbf{Step 2} constitute the initialization of the tracking process and do not need to be performed at every frame; we performed them with a subsampling rate $\mathbf{s}_{init} = 5$ meaning that tracks will only start at frames $0,5,10...$ (every 0.3s): this choice makes the initialization faster and reduces the number of wrongly initialized tracks. \textbf{Step 3} is performed track by track for every subsequent frame after initialization.

In \textbf{Step 1} we use 4 fixed overlapping regions as illustrated in Fig. 2 such that they are size $\frac{W}{2} \times H$, but detections are only kept for their central half, meaning that there are no duplicates in the global image. 

In \textbf{Step 2} we use an image of size $\frac{W}{2} \times H$ centered on each detection, using the YOLO-based bounding box center for the current frame, and the prompting bounding box is updated accordingly.

In \textbf{Step 3} segmentation and tracking are performed in an image of size  $\frac{W}{2} \times H$ centered on the SAM2-based bounding box (SAM2 yields masks but we compute their associated bounding box) of the previous frame (except at initialization). There is no reprompting and SAM2.1-Video only uses its memory bank for tracking. 
At each frame $i^*$, during the tracking of a person, a systematic comparison of its mask with all other tracked masks in the frame ($\mathcal{M}_{i^*}$) is performed \textit{(this part of \textbf{Step 3} is not displayed in Fig. 2)}; the tracking is stopped in case of very high overlap ($IoU > \mathsf{o}_{min} = 0.7$), or if the current track is enclosed for more than $95\%$ in the compared track.

When using SAM2 for segmentation and/or tracking we remove small inconsistencies of the mask, smoothen its edges and facilitate processing as well as optimize Run-Length Encoding (RLE) storage by eroding and dilating it with a kernel of size $\mathsf{k}=11$.

{
\begin{algorithm}[t]
\caption{Panoptic Person Detection, Segmentation, and Tracking}
\label{alg:panoptic}
\footnotesize
\SetKwInOut{Input}{Input}
\SetKwInOut{Output}{Output}
\SetKwBlock{Condition}{}{}

\Input{Image sequence $\{I_0,\ldots,I_{L-1}\}$}
\Input{$\mathbf{s}_{init}, \mathsf{c}_{min}^D, \mathsf{c}_{min},  \mathsf{k}_{min}, \mathsf{m}_{min}, \mathsf{p}_{min}, \mathsf{o}_{min}$}
\Output{Tracked masks $\mathcal{M} = \{\mathcal{M}_0, ..., \mathcal{M}_{L-1}\}$ with the masks of the $M_i$ tracks present in image $i$ : $\mathcal{M}_{i} = \{(\mathbf{\bar{m}}_i^{n}, \mathbf{id}^{n})\}_{n=1}^{M_i}$}

\BlankLine
\textbf{Step 1: Person Detection (YOLOv11x)}\;
\For{$i \in \{0,\mathbf{s}_{init},2\mathbf{s}_{init},\ldots\}$}{
    Split image into 4 overlapping fixed-regions: $W_R = \frac{W}{2}$, $S_R = \frac{W}{4}$, regions $R_1^i,\ldots,R_4^i$ with wrap-around\;

    \For{$k \in \{1,2,3,4\}$}{
        $R = R_k$
        
        Detect persons in $R$ : $\{\mathbf{det}_{R}^n\}_{n=1}^{D_{i,R}^{raw}}$
        $\mathbf{det}_{R}^n=[x^{\min}_{R},y^{\min}_{R},x^{\max}_{R},y^{\max}_{R}, \mathsf{c}]$\;

        Discard if 
        $x^{\min}_{R} \in [0,\frac{W_R}{4}] \cup [3\frac{W_R}{4},W_R]$\;

        Discard if $\mathsf{c} < \mathsf{c}_{min}^D$ or box size $< \mathbf{det}_{min}$

        Remap to full-image :
        $x = k S_R + x_{R}$\;

        Store $\{\mathbf{det}^1,\ldots,\mathbf{det}^{D_{i,R}}\}$\;
    }
    Store $\{\mathbf{det}^1,\ldots,\mathbf{det}^{\tilde{D}_{i}}\}$\;
}

\BlankLine
\textbf{Step 2: Segmentation and 2D Pose (SAM2.1-Image, ViTPose)}\;
\For{$i \in \{0,\mathbf{s}_{init},2\mathbf{s}_{init},\ldots\}$}{
    \For{$j \in \{1, ..., \tilde{D}_i\}$}{
        Compute center $(\tilde{x}^j,\tilde{y}^j)$ of $\mathbf{det}_{i}^j$ \;

        Extract centered-region $R = R_{i,j}$ of width $W_R$ (with wrap-around)\;

        Remap $\mathbf{det}_i^j$ into $R$ : $x_R = x - \tilde{x}_j^i - \frac{W_R}{2}$\;

        Apply SAM2.1-Image, $\mathbf{m} = \mathsf{SAM}(R_{i,j}, \mathbf{det}_i^j)$\;
        Apply ViTPose, $K = \mathsf{ViTPose}(R_{i,j}, \mathbf{det}_i^j)$\;
        $K = \{x_k,y_k,\mathsf{c}_k,\mathsf{m}_k\}_{k=1}^{17}$, $\mathsf{m_k} = 1$ if kpt $k$ in $\mathbf{m}$;

        \If{$\sum_k \mathbf{1}_{\{\mathsf{c}_k > \mathsf{c}_{min}\}} > \mathsf{k}_{min}  \mathbf{\ and\ } \sum_k \mathsf{m}_k > \mathsf{m}_{min}$ \Condition{$\mathbf{\ and\ } size(\mathbf{m}_i^j) > \mathsf{p}_{min}$}}
        {
            Remap to full image:
            $\mathbf{m}_i^j = \text{shift}(\mathbf{m}, \tilde{x}^j - \frac{W_R}{2})$\;

            Store $(\mathbf{det}_i^j,\mathbf{m}_i^j)$\;
        }
    }
    Store $\{(\mathbf{det}_i^1,\mathbf{m}_i^j),\ldots,(\mathbf{det}_i^{D_{i}},\mathbf{m}_i^{D_i})\}$\;
}

\BlankLine
\textbf{Step 3: Tracking and segmentation (SAM2.1-Video)}\;
Initialize $\forall l \in \{1,...,L-1\},  \mathcal{M}_l \leftarrow \emptyset$\;

\For{$i \in \{0,\mathbf{s}_{init},2\mathbf{s}_{init},\ldots\}$}{
    \For{$j \in \{1, ..., D_i\}$}{
    
        $(\mathbf{det}, \mathbf{m}) = (\mathbf{det}_i^1,\mathbf{m}_i^j)$
        
        \If{$\exists k$,  $\mathrm{IoU}(\mathbf{m}, \mathbf{\bar{m}}_i^k) > \mathsf{o}_{min}$}{
            continue (ignore detection)\;
        }

        Assign next $\mathbf{id}^n$, extract $R_{i,j}$ and remap $\mathbf{det}$\;
        
        $(\mathbf{S},\mathbf{\bar{m}}^i) = \mathsf{SAM2}(\mathbf{det}, R_{i,j})$ w/ $\mathbf{S}$ tracker state\;

        \For{$i^* = i+1$ \KwTo $L-1$}{
            Compute center of box associated with $\mathbf{\bar{m}}_{i^*-1}$
            
    
            $(\mathbf{S},\mathbf{\bar{m}}_{i^*}) = \mathsf{SAM2}(\mathbf{S}, R_{i^*-1,j})$\;

            \If{$size(\mathbf{\bar{m}}_{i^*}) < \mathsf{p}_{min}$ $\mathbf{\ or\ } $  $\exists k$,  $\mathrm{IoU}(\mathbf{\bar{m}}_{i^*}, \mathbf{\bar{m}}_{i^*}^k) > \mathsf{o}_{min}$}{
                break\;
            }

            Remap $\mathbf{\bar{m}}_{i^*}$ to full-image coordinates\;

            Append $(\mathbf{\bar{m}}_{i^*},\mathbf{id}^n)$ to $\mathcal{M}_{i^*}$\;
        }
    }
}

\Return{$\mathcal{M}$}\;

\end{algorithm}
}

\section{Automatic detection of interactions}

\begin{figure}
    \centering
    \includegraphics[width=0.95\linewidth]{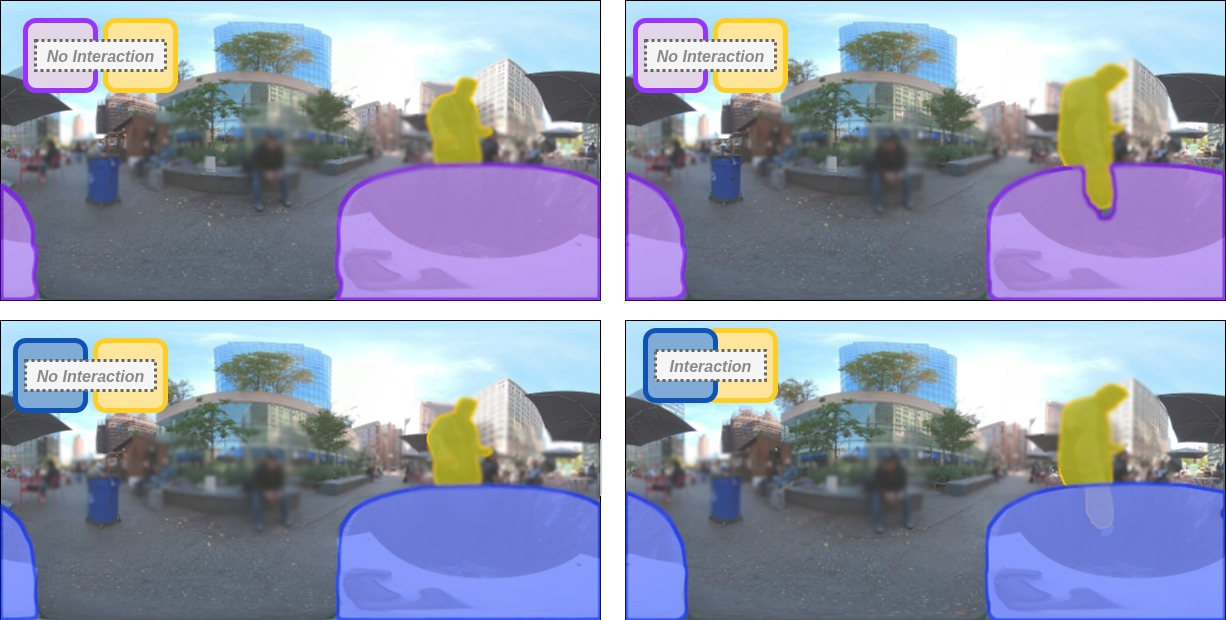}
    \caption{Using the \textcolor{blue}{\textbf{convex hull}} instead of the \textcolor{violet}{\textbf{raw}} segmented trashcan in SSUP-A to detect interactions based on masks intersections}
    \label{fig:convexhull}
\end{figure}

We automatically detected the interaction based on the intersection between a defined \textit{interaction-zone} and the SAM2-based masks of users. If they intersected on more than $\mathsf{p}^{int}_{min}$ pixels then the person was considered as interacting.

In the case of HUI360 the process was straighforward since the \textit{interaction-zone} was not moving the camera frame. In SSUP the mounting on top of the trashcans and the auto-stabilization of Insta360 cameras resulted in the \textit{interaction-zone} moving in the camera frame. To account for this \textbf{(1)} we tracked and segmented the top of trashcans using SAM2 with a single initial detection that was manually determined, \textbf{(2)} for each mask we computed its convex hull (after erosion and dilation), \textbf{(3)} we computed the intersection with the persons masks based on this convex hull as shown in \ref{fig:convexhull}.  

\section{Manual curation}

\begin{figure}
    \centering
    \includegraphics[width=0.95\linewidth]{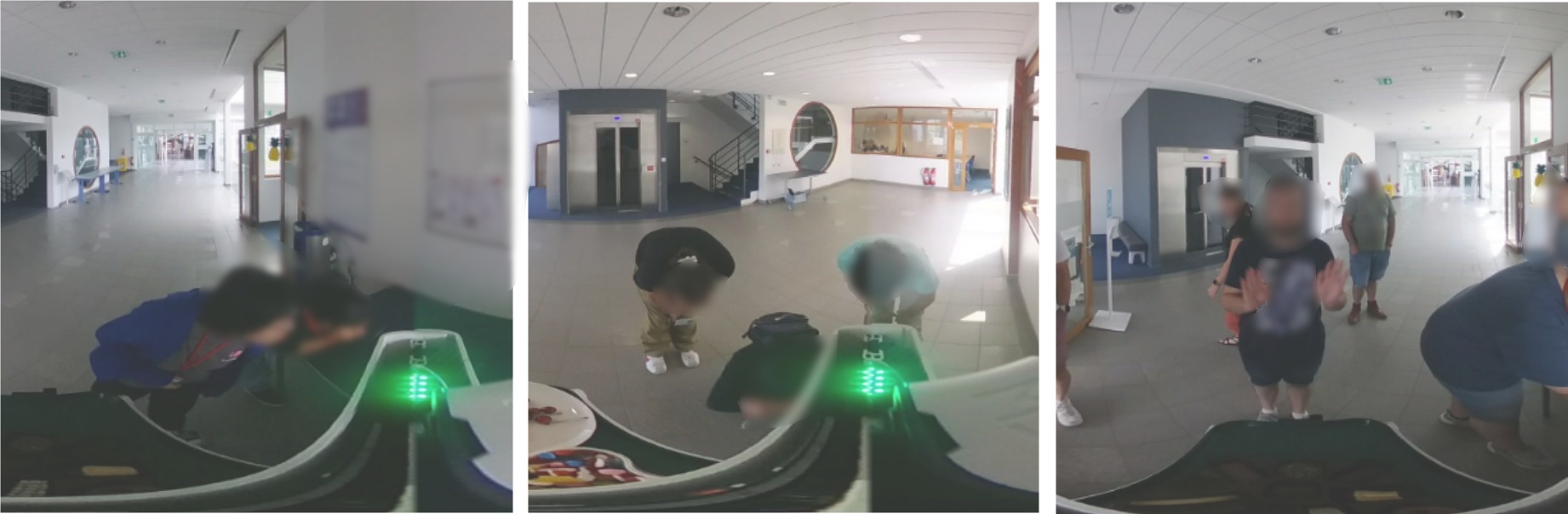}
    \caption{Discarded episodes because of behaviors attributed the novelty effect}
    \label{fig:novelty}
\end{figure}

All videos were watched and the manual curation process included the following possible actions by the annotator :
\begin{itemize}[leftmargin=+.3in]
    \item Discard a full episode
    \item Discard a track (will not be exported but other tracks in the episode will be exported)
    \item Split a track
    \item Merge two tracks into one
    \item Adjust the \textit{current interaction} label for a track during a segment of time
\end{itemize}

These actions were performed with a custom dedicated tool (\ref{fig:correctiontool}) that is made available in the code as part of this submission. We did not directly corrected the masks or the extracted keypoints, in case of failure the tracks were fully discarded. 

Among the reason to fully discard an episode was the important presence of "abnormal" behaviors, that could be largely attributed to the novelty and curiosity inspired by the robot (\ref{fig:novelty}) since the goal was to capture every day interaction as they would occur in a situation of long-term deployment.

\begin{figure}
    \centering
    \includegraphics[width=1.0\linewidth]{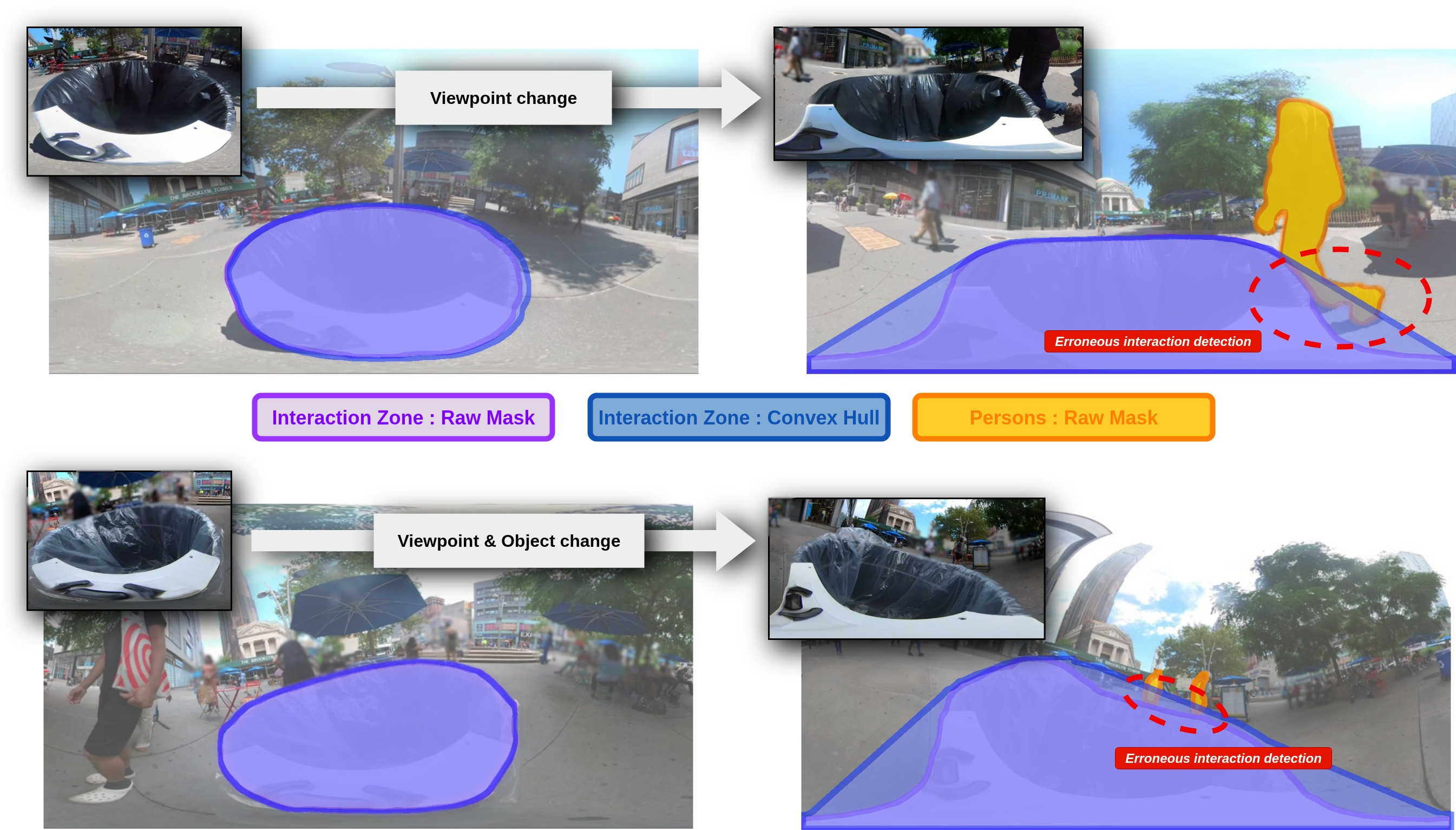}
    \caption{False positives in SSUP-A when using intersection of persons masks with \textit{interaction zone} convex hull under skewed viewpoint}
    \label{fig:problemhull}
\end{figure}

\begin{figure}
    \centering
    \includegraphics[width=0.95\linewidth]{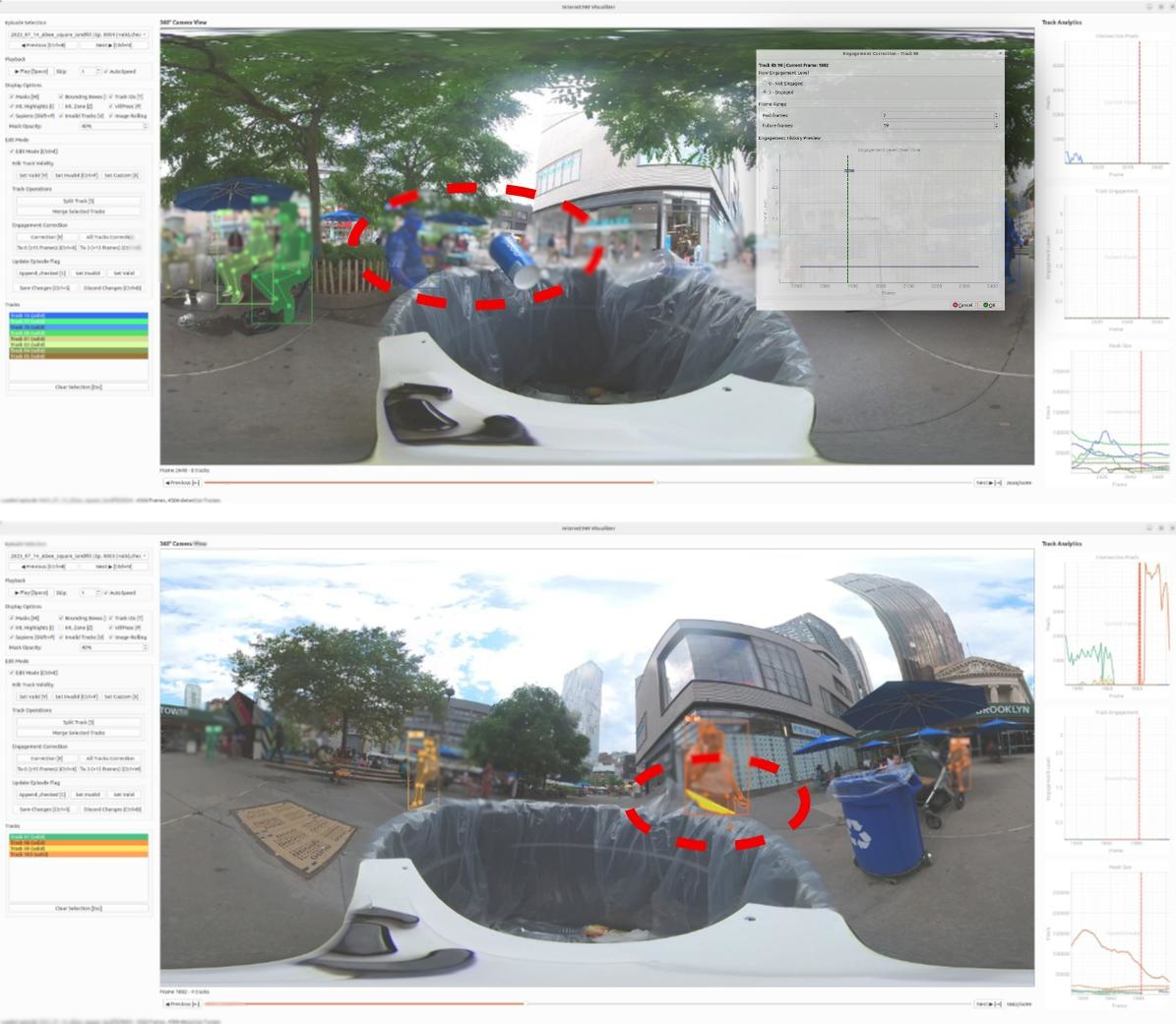}
    \caption{Example of corrections made to automatic interaction detection in SSUP-A. \textit{Screenshots of the visualization/correction tool}.}
    \label{fig:correctiontool}
\end{figure}

The \textit{current interaction} label mostly had to be adjusted in tracks of SSUP-A : \textbf{(1)} because of false negatives : since getting rid of garbage is semantically the same action and comes from the same intent whether it was done by placing an object in the can or by throwing it, we consistently considered as interacting people that were throwing away stuff and manually added interaction labels to those doing it without having their mask intersecting with the \textit{interaction zone} and \textbf{(2)} because of false positives that were the result of using the convex hull of the interaction zone in situations were they appeared non-convex in the image (because of the camera position and auto-stabilization, or because of external factors like the trash bags moving) as shown in \ref{fig:problemhull}.


\section{Precision on Problem statement}

\begin{figure*}
    \centering
    \includegraphics[width=\linewidth]{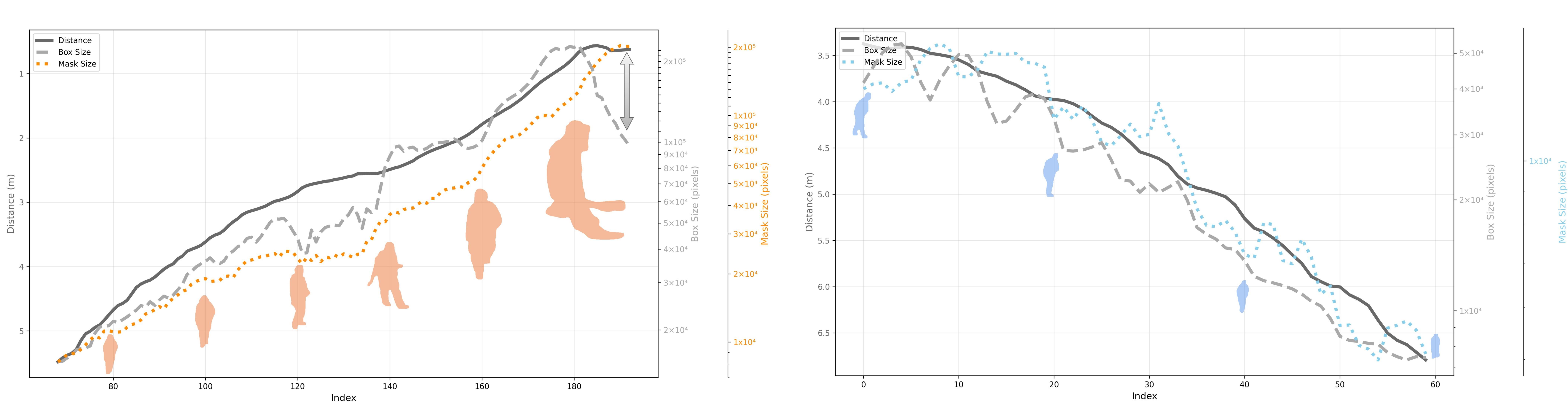}
    \caption{Using the mask size as a proxy for distance. Note that the scales used for mask and box sizes (in pixels) are logarithmic. All the illustrative overlayed masks have the same scale.}
    \label{fig:proxy}
\end{figure*}

We used the mask size as a proxy for distance when aligning the negative tracks during sampling, for those tracks $\mathbf{T_0}$ is define as the moment where they appear with the biggest mask size. We show in \ref{fig:proxy} the obvious correlation between mask size (in the full 360 image) and distance (measured with an additional RGBD camera), we also compare it to the box size. The box size also appears correlated to the distance but we found it to be less stable (more subject to bias depending on position and movement whereas the mask size is stable except in the presence of severe occlusions, and these cases are usually filtered out during sampling as explained in \ref{DetailsSup}). 

\section{Implementation details}\label{DetailsSup}

The sampling procedure to generate $\mathcal{X} = \{X_1,...,X_N\}$ with $X_i \in \mathbb{R}^{T\times D}$  and $\mathcal{Y} = \{y_1,...,y_N\}$ with $y_i\in\{0,1\}$ is introduced in Sec. 3.5 and Sec. 4.2 and detailed in  \ref{alg:track_sampling}. It involves variables $\mathbf{T^{org}}$ (input duration in frames before eventual subsampling), $\mathbf{s}$ (subsampling rate), $T$ input duration in frame after subsampling, $\mathbf{T_{POS}}$ (cutoff for positive labels) and $\mathbf{T_{CUT}}$ (cutoff to discard tracks after interaction or when they are moving away).

\section{Additional feature sets in baselines}

We report the results obtained with different feature sets in \ref{tab:cross_detailed} and \ref{tab:detailedadv} for the cross-dataset baselines and for the various advance detection time baselines. In Tab. 3 and Tab. 4 all the results are reported using only $\mathcal{D}_3$ which is in many cases the best choice.

In \ref{tab:cross_detailed}  each model is trained on its native dataset, and subsequently evaluated in a in-dataset or cross-dataset transfer scenario. We compare feature set $\mathcal{D}_3$ (mask size, box and ViTPose keypoints), $\mathcal{D}_6$ (mask size, box and head + shoulders keypoints) and $\mathcal{D}_7$ (mask size and ViTPose keypoints) and $\mathbf{T_{ADV}}=15$.

In \ref{tab:detailedadv}, classifiers are trained and evaluated with the same $\mathbf{T_{ADV}} \in \{5,10,15,20,25,30\}$, corresponding to a forecasting of 0.33, 0.66, 1.0, 1.33, 1.66 or 2.0 seconds respectively. Note that for consistency all of them are only trained on a set of tracks from from HUI360 Train and evaluated on set of tracks from HUI360 Test that are both long enough to allow sampling with $\mathbf{T_{ADV}}$ up to $30$.

\begin{table}[H]
\centering
\fontsize{8}{10}\selectfont
\setlength{\tabcolsep}{5pt} 
\renewcommand{\arraystretch}{1.0} 
\begin{tabular}{cccccccc}
\toprule
{} & {} & \multicolumn{3}{c}{HUI360 Test} & \multicolumn{3}{c}{SSUP-A Test} \\
\cmidrule(lr){3-5}\cmidrule(lr){6-8}
{} & & {} \textbf{RF} & \textbf{MLP} & \textbf{LSTM} & \textbf{RF} & \textbf{MLP} & \textbf{LSTM} \\
\midrule
\multirow{3}{*}{\makecell[t]{HUI360\\ Train}} & $\mathcal{D}_3$ & \textbf{0.81} & 0.86 & \textbf{0.91} & \textbf{0.83} & \textbf{0.79} & 0.84 \\
 & $\mathcal{D}_6$ & \cellcolor{gray!20}{\textbf{0.81}} & \cellcolor{gray!20}{0.86} & \cellcolor{gray!20}{\textbf{0.91}} & \cellcolor{gray!20}{\textbf{0.83}} & \cellcolor{gray!20}{0.75} & \cellcolor{gray!20}{0.82} \\
 & $\mathcal{D}_7$ & \cellcolor{GreenYellow!20}{0.80} & \cellcolor{GreenYellow!20}{\textbf{0.90}} & \cellcolor{GreenYellow!20}{0.89} & \cellcolor{GreenYellow!20}{0.81} & \cellcolor{GreenYellow!20}{\textbf{0.79}} & \cellcolor{GreenYellow!20}{\textbf{0.85}} \\
 \midrule
\multirow{3}{*}{\makecell[t]{SSUP-A\\ Train}} & $\mathcal{D}_3$ & 0.76 & 0.81 & 0.84 & \textbf{0.86} & 0.85 & 0.88 \\
 & $\mathcal{D}_6$ & \cellcolor{gray!20}{\textbf{0.77}} & \cellcolor{gray!20}{0.79} & \cellcolor{gray!20}{0.83} & \cellcolor{gray!20}{0.85} & \cellcolor{gray!20}{0.85} & \cellcolor{gray!20}{0.87} \\
 & $\mathcal{D}_7$ & \cellcolor{GreenYellow!20}{0.76} & \cellcolor{GreenYellow!20}{\textbf{0.82}} & \cellcolor{GreenYellow!20}{\textbf{0.86}} & \cellcolor{GreenYellow!20}{\textbf{0.86}} & \cellcolor{GreenYellow!20}{\textbf{0.86}} & \cellcolor{GreenYellow!20}{\textbf{0.89}} \\
\bottomrule
\end{tabular}
\caption{AUC of RF, MLP and LSTM. Best feature set per classifier on each pair of train-test set in \textbf{bold}.}
\label{tab:cross_detailed}
\end{table}

\begin{table}[H]
\caption{AUC of RF, MLP and LSTM classifier at different advance detection thresholds $\mathbf{T_{ADV}}$ and with different feature sets $\mathcal{D}$. Best feature set per classifier and per $\mathbf{T_{ADV}}$ in \textbf{bold}}
\centering
\fontsize{8}{10}\selectfont
\begin{tabular}{cccccccc}
\toprule
\multicolumn{2}{c}{$\mathbf{T_{ADV}}$} & 5 & 10 & 15 & 20 & 25 & 30 \\
\midrule
\multirow{3}{*}{\textbf{RF}} & $\mathcal{D}_3$ & \textbf{0.90} & \textbf{0.84} & \textbf{0.81} & \textbf{0.78} & \textbf{0.73} & 0.71 \\
 & $\mathcal{D}_6$ & \cellcolor{gray!20}{\textbf{0.90}} & \cellcolor{gray!20}{0.83} & \cellcolor{gray!20}{0.78} & \cellcolor{gray!20}{0.76} & \cellcolor{gray!20}{0.71} & \cellcolor{gray!20}{0.71} \\
 & $\mathcal{D}_7$ & \cellcolor{GreenYellow!20}{0.88} & \cellcolor{GreenYellow!20}{\textbf{0.84}} & \cellcolor{GreenYellow!20}{0.80} & \cellcolor{GreenYellow!20}{0.76} & \cellcolor{GreenYellow!20}{\textbf{0.73}} & \cellcolor{GreenYellow!20}{\textbf{0.72}} \\
 \midrule
\multirow{3}{*}{\textbf{MLP}} & $\mathcal{D}_3$ & \textbf{0.96} & 0.92 & \textbf{0.89} & \textbf{0.86} & \textbf{0.84} & 0.76 \\
 & $\mathcal{D}_6$ & \cellcolor{gray!20}{\textbf{0.96}} & \cellcolor{gray!20}{\textbf{0.94}} & \cellcolor{gray!20}{0.88} & \cellcolor{gray!20}{0.84} & \cellcolor{gray!20}{0.81} & \cellcolor{gray!20}{\textbf{0.77}} \\
 & $\mathcal{D}_7$ & \cellcolor{GreenYellow!20}{\textbf{0.96}} & \cellcolor{GreenYellow!20}{0.93} & \cellcolor{GreenYellow!20}{\textbf{0.89}} & \cellcolor{GreenYellow!20}{0.84} & \cellcolor{GreenYellow!20}{0.83} & \cellcolor{GreenYellow!20}{0.76} \\
 \midrule
\multirow{3}{*}{\textbf{LSTM}} & $\mathcal{D}_3$ & \textbf{0.97} & \textbf{0.94} & 0.89 & \textbf{0.87} & \textbf{0.84} & \textbf{0.77} \\
 & $\mathcal{D}_6$ & \cellcolor{gray!20}{\textbf{0.97}} & \cellcolor{gray!20}{\textbf{0.94}} & \cellcolor{gray!20}{\textbf{0.90}} & \cellcolor{gray!20}{0.83} & \cellcolor{gray!20}{0.82} & \cellcolor{gray!20}{0.74} \\
 & $\mathcal{D}_7$ & \cellcolor{GreenYellow!20}{0.95} & \cellcolor{GreenYellow!20}{0.92} & \cellcolor{GreenYellow!20}{0.88} & \cellcolor{GreenYellow!20}{0.83} & \cellcolor{GreenYellow!20}{0.81} & \cellcolor{GreenYellow!20}{\textbf{0.77}} \\
\bottomrule
\end{tabular}
\label{tab:detailedadv}
\end{table}

\section{Detailed results}




\begin{figure*}[t]
\centering
\begin{minipage}[t]{0.48\textwidth}
  \centering
  \includegraphics[width=\linewidth]{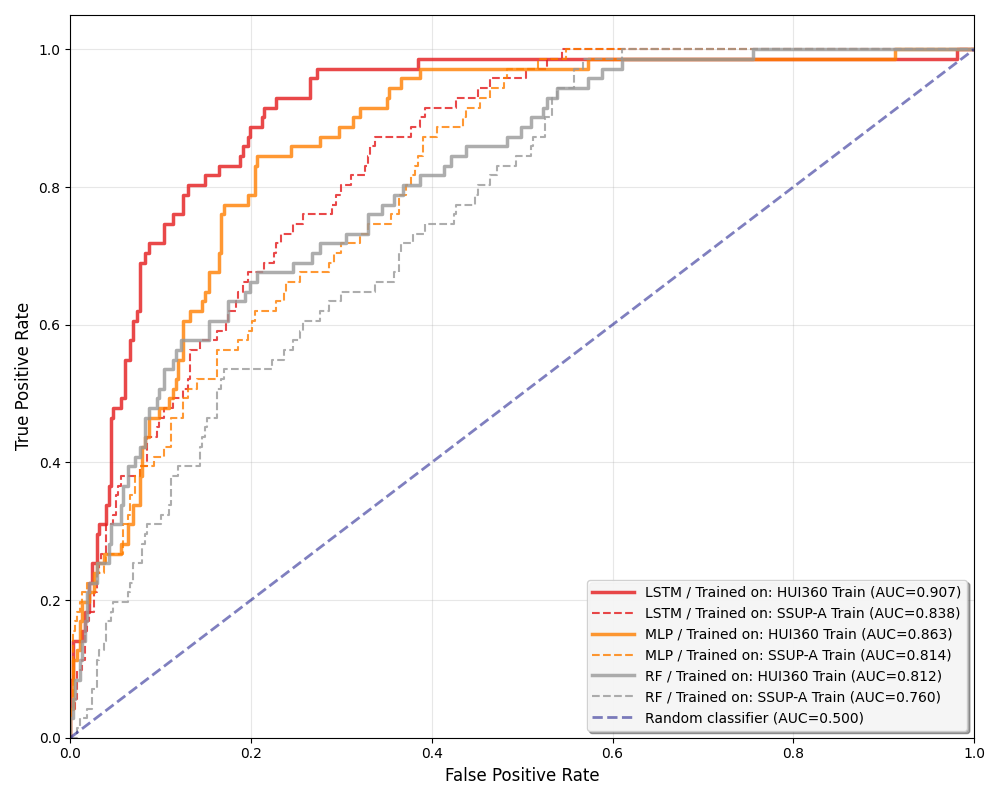}
  \par\small (a) Evaluated on HUI360 Test
\end{minipage}\hfill
\begin{minipage}[t]{0.48\textwidth}
  \centering
  \includegraphics[width=\linewidth]{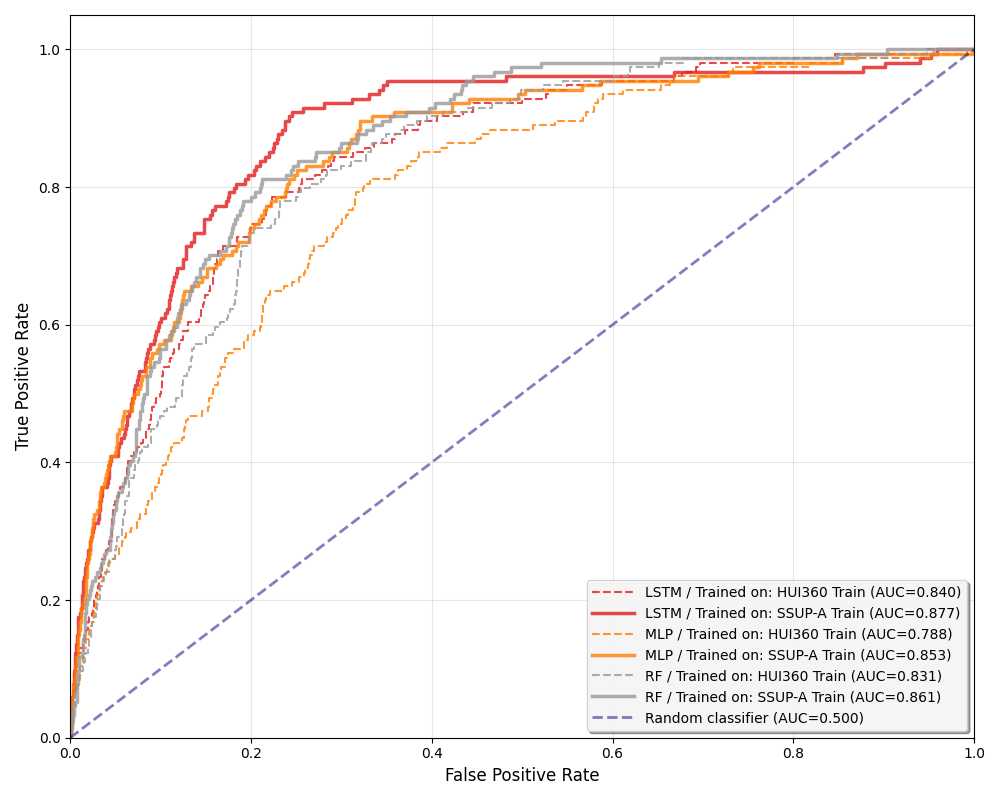}
  \par\small (b) Evaluated on SSUP-A Test
\end{minipage}

    \caption{ROC curves of classifiers (RF, MLP, LSTM) for cross-dataset evaluation using $\mathcal{D}_3$ and $\mathbf{T_{ADV}} = 15$. Classifiers are trained on HUI360 Train or SSUP-A Train, and evaluated on HUI360 Test (left) or SSUP-A Test (right). In HUI360 Test, $TP=71$ and $TN=377$, in SSUP-A Test, $TP=154$ and $TN=5080$.}
\label{fig:crossroc}
\vspace{18pt}
\end{figure*}

We report the ROC curves illustrating the results of Tab. 3 (cross-dataset baselines), in \ref{fig:crossroc}.
Under reasonable False Postive Rate, the LSTM consistently outperforms the MLP and RF baselines.

\section{Interaction anticipation examples}

\begin{figure*}
    \centering
    \includegraphics[width=0.95\linewidth]{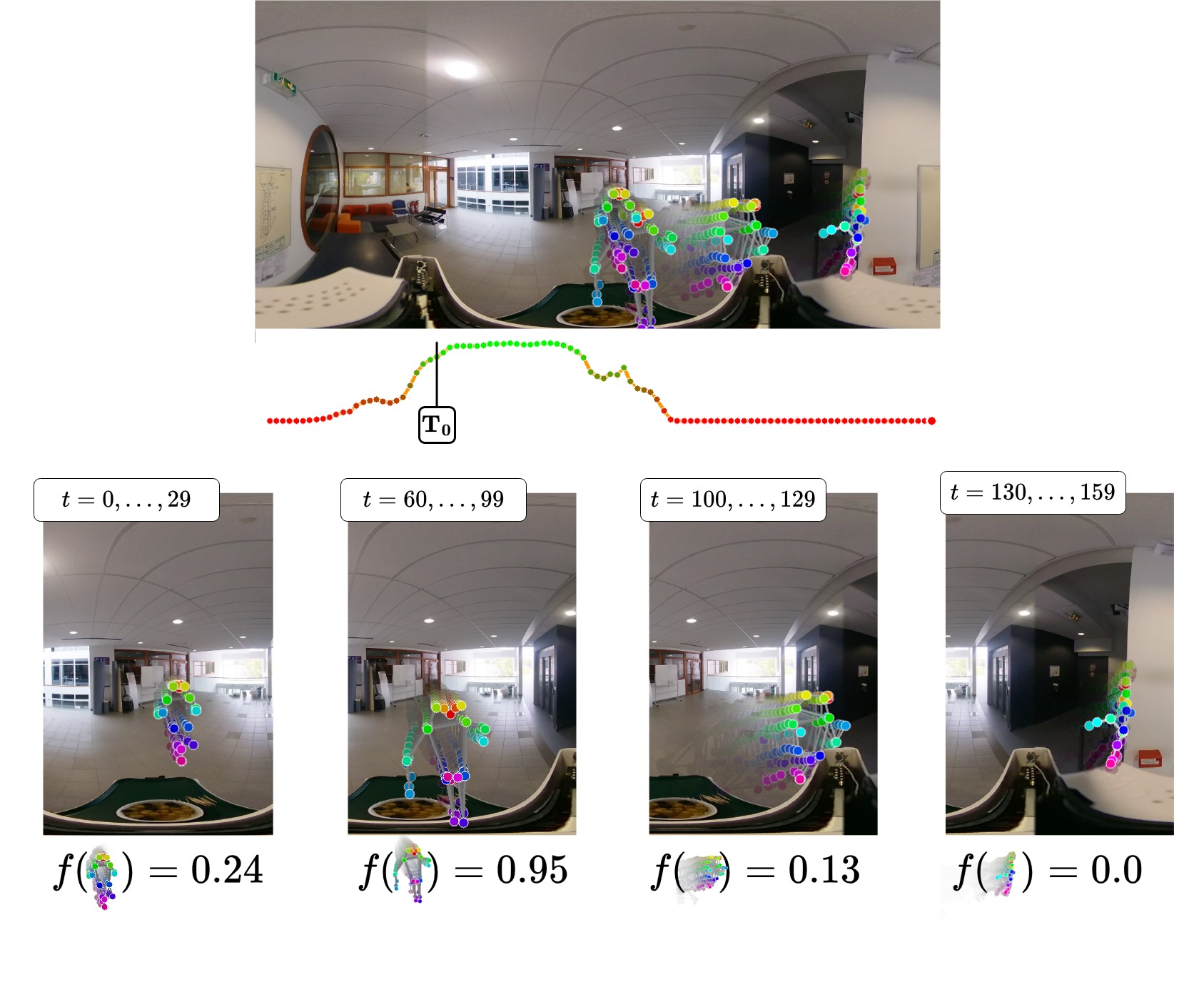}
    \caption{Predictions of an MLP classifier on 4 different segments of the same track}
    \label{fig:baselineprediction}
\end{figure*}

We show an example of prediction with an MLP classifier on 4 segments of a track sampled at different times in \ref{fig:baselineprediction} and refer to the supplementary video for additional predictions.

{
\begin{algorithm}[t]
\caption{Track Sampling}
\label{alg:track_sampling}
\footnotesize
\SetKwInOut{Input}{Input}
\SetKwInOut{Output}{Output}
\SetKwBlock{Condition}{}{}

\Input{Tracks $\mathcal{T} = \{\mathcal{T}_1,...,\mathcal{T}_{N_{raw}}\}$ with $\mathcal{T}_i = \{x_{i,j}\}_{j=0}^{T_i-1}$ where $x_{i,j} \in \mathbb{R}^{D}$ are the features of track $i$ at time $t_j$ (with contiguous time: $t_{j+1} - t_j = \frac{1}{f}$ seconds)}
\Input{\textit{Current} interaction labels $\{a_{i,j}\}_{j=0}^{T_i-1}$, $a_{i,j}\in \{0,1\}$}

\Input{$\mathbf{T^{org}}$, $\mathbf{s}$, $T$, $\mathbf{T_{CUT}}$, $\mathbf{T_{POS}}$, $\mathsf{k}_{min}$, $\mathsf{c}_{min}$}
\Output{$\mathcal{X}, \mathcal{Y}$: one segment and \textit{future} interaction label per track}

\BlankLine
\textbf{Step 1: Determine onset per track}\;

\For{$i \in \{1,...,N_{raw}\}$}{
    Compute onset time:
    \[
        \mathbf{T}_\mathbf{0}^i = 
        \begin{cases}
        \mathrm{min}_j\{j,\ a_{i,j}=1\}, & \text{if exists} \\
         \mathrm{argmax}_j\{m_{i,j}\}, & \text{otherwise}
        \end{cases}
    \]

    Where $m_{i,j}$ is the mask size of track $i$ at time $t_j$
    
    Store $\mathbf{T}_\mathbf{0}^i$
}

\BlankLine
\textbf{Step 2: Per-frame validity filtering}\;

\For{$i \in \{1,...,N_{raw}\}$}{
    \For{$j = 0$ \KwTo $T_i-1$}{
    Compute a validity flag for each frame of each track
    \[
        \mathsf{v}_{i,j} = 
        \begin{cases}
        1, & \text{if} \sum_k \mathbf{1}_{\{\mathsf{c}_{i,j,k} > \mathsf{c}_{min}\}} > \mathsf{k}_{min} \\
        0, & \text{otherwise}
        \end{cases}
    \]
    Where $\mathsf{c}_{i,j,k}$ is the confidence associated to the $k$-th keypoints of the track $i$ at time $t_j$
    }
    Store $\{\mathsf{v}_{i,j}\}_{j=0}^{T_i-1}$
}



\BlankLine
\textbf{Step 3: Extract all valid contiguous segments}\;

\For{$i \in \{1,...,N_{raw}\}$}{
    Initialize segments $\mathcal{S}^i \leftarrow \emptyset$ and labels $\mathcal{L}^i \leftarrow \emptyset$\;
    \If{$\mathbf{T}_\mathbf{0}^i - \mathbf{T_{CUT}} < \mathbf{T^{org}}$}{
        continue (ignore track)\;
    }
    \For{$b = 0$ \KwTo $\mathbf{T}_\mathbf{0}^i - \mathbf{T_{CUT}} - \mathbf{T^{org}} - 1$}{
        \If{$\mathrm{all}(\{\mathsf{v}_{i,j} = 1\}_{j=b}^{b+\mathbf{T^{org}}})$}{
            \[
                S_b^i = [x_{i,b},\ldots,x_{i,b+T}] \in \mathbb{R}^{\mathbf{T^{org}}\times D}
            \]
            \[
                l_b^i = 
                \begin{cases}
                1, & \text{if}\ \exists j, a_{i,j} = 1\ \mathbf{and}\ b+\mathbf{T^{org}} \ge \mathbf{T}_\mathbf{0}^i - \mathbf{T_{POS}}\\
                0, & \text{otherwise}
                \end{cases}
            \]
            
            Append $S_b^i$ to $\mathcal{S}^i$ and $l_b^i$ to $\mathcal{L}^i$\;
        }
    }
    Store $(\mathcal{S}^i,\mathcal{L}^i) = (\{S_s^i\}_{s=1}^{N_i},\{l_s^i\}_{s=1}^{N_i})$
}

\BlankLine
\textbf{Step 4: Final sampling (one segment per track)}\;

Let $\{(\mathcal{S}^i,\mathcal{L}^i)\}_{i=1}^{N}$ the pairs of segments-labels for the $N$ valid tracks such that $\mathcal{S}^i \ne \emptyset$

Initialize $\mathcal{X} \leftarrow \emptyset$ and labels $\mathcal{Y} \leftarrow \emptyset$\;

\For{$i \in \{1,...,N\}$}{

    Randomly$^{\cross[.4pt]}$ sample $j^* \in \{1,...,N_i\}$

    \If{$\mathbf{s} \ne 1$}{

    Subsample with rate $\mathbf{s}$, $T = \ceil*{\frac{\mathbf{T^{org}}}{\mathbf{s}}}$

    Given $S^i_{j^*} = [x_{i,0}, x_{i,1},,..., x_{i,\mathbf{T^{org}}}] \in \mathbb{R}^{\mathbf{T^{org}}\times D}$
    
    \[
    S^i_{j^*}\leftarrow [x_{i,0}, x_{i,\mathbf{s}}, x_{i,2\mathbf{s}},...] \in \mathbb{R}^{T\times D}
    \]
    
    }
    
    Append $X_{i} = S^i_{j^*} \in \mathbb{R}^{T\times D}$ to $\mathcal{X}$
    
    Append $y_{i} = l^i_{j^*} \in \{0,1\}$ to $\mathcal{Y}$
}

\Return{$\mathcal{X}, \mathcal{Y}$}\;

\vspace{3pt}

\rule{0.45\textwidth}{1pt}

\vspace{3pt}

\scriptsize
\cross[.4pt] In practice to maximize the number of positive samples, we choose $j^*$ such that $y_{j^*} = 1$ when possible for the track 

\vspace{3pt}
\end{algorithm}
}

\end{document}